\documentclass{new_tlp}
\usepackage{mathptmx}

\usepackage[utf8]{inputenc}
\newcommand{\trash}[1]{}

\newcommand{\shortversion}[1]{}

\usepackage[USenglish]{babel}

\usepackage[hyphens,spaces]{url}
\usepackage{hyperref}
\usepackage[]{xcolor}

\newcommand{\tuplecolor}[1]{\textcolor{#1}}
\newcommand{\inputPredColor}{orange!55!red}

\newcommand{\statePredColor}{green!62!black}
\newcommand{\specialPredColor}{red!62!black}

\newcommand{\Tab}[1]{\ensuremath{\text{C-Tabs}}}

\usepackage{caption}
\usepackage{graphicx}

\usepackage{amsmath,amsfonts,amssymb}
\usepackage{booktabs}
\usepackage{alltt}
\usepackage{microtype}

\usepackage{multirow}

\makeatletter
\newcommand\footnoteref[1]{\protected@xdef\@thefnmark{\ref{#1}}\@footnotemark}
\makeatother

\newcommand{\cid}[1]{\ensuremath{#1}}

\usepackage[ruled,vlined,linesnumbered]{algorithm2e}
\newcommand*{\algorithmcfnameold}{Algorithm}
\newcommand*{\algorithmcfnamenew}{Algorithm}
\renewcommand*{\algorithmcfname}{\algorithmcfnameold}
\SetKwInput{KwData}{In}
\SetKwInput{KwResult}{Out}
\SetAlFnt{\small}
\SetAlCapFnt{\small}
\SetAlCapNameFnt{\small}
\SetAlCapHSkip{0pt}
\SetEndCharOfAlgoLine{}
\IncMargin{-0.4em}
\makeatletter
\newcommand{\algorithmfootnote}[2][\footnotesize]{
  \let\old@algocf@finish\@algocf@finish
  \def\@algocf@finish{\old@algocf@finish
    \leavevmode\rlap{\begin{minipage}{\linewidth}
    #1#2
    \end{minipage}}
  }
}
\makeatother

\usepackage{rotating}
\usepackage{xspace}

\newcommand{\SB}{\{\,}%
\newcommand{\SM}{\;{|}\;}%
\newcommand{\SE}{\,\}}%

\newcommand{\sharpP}{\#P\xspace}

\newcommand{\eqdef}{\ensuremath{\,\mathrel{\mathop:}=}}
\newcommand{\TTT}{\mathcal{T}}
\newcommand{\Card}[1]{|#1|}
\let\phi=\varphi
\let\epsilon=\varepsilon

\newcommand{\SAT}{\textsc{Sat}\xspace}%
\newcommand{\MSAT}{\textsc{MaxSat}\xspace}%
\newcommand{\cSAT}{\textsc{\#Sat}\xspace}%
\newcommand{\cTCOL}{\textsc{\#$o$-Col}\xspace}%
\newcommand{\VC}{\textsc{MinVC}\xspace}%
\newcommand{\IDS}{\textsc{MinIDS}\xspace}%
\DeclareMathOperator{\tw}{tw}
\DeclareMathOperator{\dom}{dom}
\DeclareMathOperator{\attr}{cols}
\DeclareMathOperator{\ass}{ass}
\DeclareMathOperator{\width}{width}

\newcommand{\dpdb}{{\small\textsf{dpdb}}\xspace}
\newcommand{\dpdbold}{{\small\textsf{dpdb pg9}}\xspace}

\newcommand{\gpusatnu}{{{gpusat2}}\xspace}

\newcommand{\gpusatone}{{{gpusat1}}\xspace}

\newcommand{\instances}[1]{\texttt{#1}}

\newcommand{\etal}{et~al.\@\xspace}

\newcommand{\algo}[1]{\ensuremath{\mathsf{#1}}}

\newcommand{\tab}[1]{\ensuremath{\tau_{#1}}}

\newcommand{\CTabs}[1]{\ensuremath{\text{C-Tabs}}}

\DeclareMathOperator{\var}{var}
\DeclareMathOperator{\tvar}{tvar}

\DeclareMathOperator{\type}{type}
\newcommand{\intr}{\textit{intr}}
\newcommand{\leaf}{\textit{leaf}}
\newcommand{\rem}{\textit{rem}}
\newcommand{\join}{\textit{join}}

\newtheorem{example}{Example} %
\newtheorem{EXa}{Example} %

\renewenvironment{example}{\begin{EXa}}{\hfill\ensuremath{\blacksquare}\end{EXa}}

\title[Exploiting Database Management Systems and Treewidth for
Counting]{Exploiting Database Management Systems and Treewidth for
  Counting\footnote{This is an extended version of a
    paper~\cite{FichteEtAl20} that appeared in the Proceedings of the
    22nd International Symposium on Practical Aspects of Declarative
    Languages (PADL'20).
}
}%

\author[Fichte~\protect\etal]
{Johannes K. Fichte\\%
	UC Berkeley, USA\\
	\email{johannes.fichte@berkeley.edu}%
	\and Markus Hecher\\
	TU Wien, Austria\\
	\email{hecher@dbai.tuwien.ac.at}%
	\and Patrick Thier\\
	TU Wien, Austria\\
	\email{thier@dbai.tuwien.ac.at}%
	\and Stefan Woltran\\
	TU Wien, Austria\\
	\email{woltran@dbai.tuwien.ac.at}}%

\jdate{March 2003}
\pubyear{2003}
\pagerange{\pageref{firstpage}--\pageref{lastpage}}
\doi{S1471068401001193}

\begin{document}
\maketitle%

\begin{abstract}%
Bounded treewidth is one of the most cited combinatorial invariants in the literature. It  was also applied  for solving several counting problems efficiently. 
A canonical counting problem is \cSAT, which asks to count the satisfying assignments of a Boolean formula. Recent work shows that benchmarking instances for \cSAT often have reasonably small treewidth. %
This paper deals with counting problems for instances of small treewidth. We introduce a general framework to solve counting questions based on state-of-the-art database management systems (DBMSs). Our framework takes explicitly advantage of small treewidth by solving instances using dynamic programming (DP) on tree decompositions (TD). Therefore, we implement the concept of DP into a DBMS (PostgreSQL), since DP algorithms are already often given in terms of table manipulations in theory. This allows for elegant specifications of DP algorithms and the use of SQL to manipulate records and tables, which gives us a natural approach to bring DP algorithms into practice. To the best of our knowledge, we present the first
approach to employ a DBMS for algorithms on TDs. A key advantage of our approach is that DBMSs naturally allow for dealing with huge tables with a limited amount of main memory (RAM). %
\end{abstract}

\section{Introduction}

\noindent Counting solutions is a well-known task in mathematics, computer
science, and other
areas \cite{ChakrabortyMeelVardi16a,DomshlakHoffmann07a,GomesKautzSabharwalSelman08a,SangBeameKautz05a}.
In combinatorics, for instance, one characterizes the
number of solutions to problems by means of mathematical
expressions,~e.g., generating
functions~\cite{DoubiletRotaStanley1972}.
One particular counting problem, namely \emph{model counting} (\cSAT) asks to output the number
of solutions of a given Boolean formula.
While we stay in the realm of model counting,
the findings of this work are also relevant for
answer set programming.
This is particularly true for tight programs (using, e.g., Clark's completion~\cite{Clark77}),
but also interesting for applications of quantitative reasoning, solved by 
programs that are compiled to \SAT
with the help of tools like lp2sat~\cite{Janhunen06} or lp2acyc~\cite{BomansonEtAl16}.

Model counting and variants thereof have already
been applied for solving a variety of real-world applications and
questions in modern society related to
reasoning~\cite{ChakrabortyMeelVardi16a,ChoiBroeckDarwiche15a,MeelEtAl17a,XueChoiDarwiche12a}.
Such problems are typically considered very hard, since \cSAT is
complete for the class
\sharpP~\cite{BacchusDalmaoPitassi03,Roth96},~i.e., one can simulate
any problem of the polynomial hierarchy with polynomially many
calls~\cite{Toda91} to a~\cSAT solver.  Taming this high complexity is
possible with techniques from parameterized
complexity~\cite{CyganEtAl15}.  In fact, many of the publicly
available~\cSAT instances show good structural properties after using
regular preprocessors like pmc~\cite{LagniezMarquis14a},
see~\cite{FichteEtAl18c,FichteHecherZisser19a}.
By good structural properties, we mean that graph representations of these instances
have reasonably small \emph{treewidth}.
The measure treewidth is a structural parameter of graphs
which models the closeness of the graph of being a tree.
Treewidth  is one of the most cited %
combinatorial invariants studied in parameterized complexity~\cite{CyganEtAl15} and was subject to algorithmics competitions~\cite{DellKomusiewiczTalmon18a}.

The observation, stated above, that various recent problem instances for \#SAT have small treewidth,
leads to the question whether a general framework that leverages treewidth is possible for counting problems.
The general idea to develop such frameworks is indeed not new, since
there are (a)~specialized solvers such as dynQBF, gpuSAT, and
fvs-pace~\cite{CharwatWoltran17,FichteHecherZisser19a,KiljanPilipczuk18}
as well as (b)~general systems that exploit treewidth like
D-FLAT~\cite{BliemEtAl16}, Jatatosk~\cite{BannachBerndt19}, and
sequoia~\cite{LangerEtAl12}.
Some of these systems explicitly use \emph{dynamic programming (DP)} 
to directly exploit treewidth
by means of so-called \emph{tree decompositions (TDs)},
whereas others provide some kind of declarative layer to model
the problem (and perform decomposition and DP internally).
In this work, we solve (counting) problems by means of explicitly specified DP algorithms,
where essential parts of the 
DP algorithm are specified in form of SQL {\ttfamily SELECT} queries.
The actual run of the DP algorithm is then delegated to our system~\dpdb,
which employs \emph{database management systems (DBMSs)}~\cite{Ullman89}.
This has not only the advantage of naturally describing and manipulating the tables
that are obtained during DP, but also allows \dpdb to benefit from decades of
database technology in form of the capability to deal with huge tables
using a limited amount of main memory (RAM),
dedicated database joins, 
query optimization, and data-dependent execution plans.
Compared to other generic DP systems like D-FLAT~\cite{BliemEtAl16}, our system \dpdb
uses relational algebra (SQL) for specifying DP algorithms,
which is even competitive with specialized systems for model counting,
and therefore applicable beyond rapid prototyping.
\paragraph{Contribution.}  
We implement a system \dpdb for solving counting problems based on dynamic programming on tree decompositions,
and present the following contributions.
(i)~Our system \dpdb uses database management systems to handle table operations needed
for performing dynamic programming efficiently. The system \dpdb is written in Python and employs PostgreSQL as DBMS,
but can work with other DBMSs easily.
(ii)~The architecture of \dpdb allows to solve general problems of bounded treewidth that can be solved
by means of table operations (in form of relational algebra and SQL) on tree decompositions. As a result, \dpdb is a generalized framework 
for dynamic programming on tree decompositions, where one only needs to specify the essential and problem-specific parts of dynamic programming
in order to solve (counting) problems.
(iii)~Finally, we show how to solve the canonical problem \cSAT with the help of \dpdb, 
where it seems that the architecture of \dpdb
is particularly well-suited. In more detail, we compare the runtime of
our system with state-of-the-art model counters. We observe a competitive behavior
and promising indications for future work.

\paragraph{Prior Work}
This is an extended version of a paper~\cite{FichteEtAl20} that appeared at the
22nd International Symposium on Practical Aspects of Declarative Languages.
The new material includes improved and extended examples
as well as a detailed description of our DP algorithms and how
these algorithms can be implemented for the system \dpdb.
Further, we added two new DP algorithms for the additional problems \MSAT and \IDS,
to demonstrate how to use \dpdb:
The problem \MSAT is similar to \SAT, but consists of hard clauses that need to be satisfied
as well as soft clauses. The goal of \MSAT is to compute the maximum number of soft clauses
that can be satisfied using only assignments that also satisfy all the hard clauses.
Further, problem \IDS is a popular graph problem that aims for computing for a given graph, a set of vertices (called independent dominating set) such that there is no edge between these vertices
and all the other vertices of the graph have an edge to at least one of the vertices in this set.
Both problems can be easily extended to counting, where we require to compute the number of witnessing
solutions.
Finally, we added new experimental results, where we used the most recent version~12 of PostgreSQL
as the underlying  database management system, which operated on a ramdisk drive.

\section{Preliminaries}%

We assume familiarity with the terminology of graphs and trees.
For details, we refer to the literature and standard textbooks~\cite{Diestel12}.

\subsection{Boolean Satisfiability}
  We define Boolean formulas and their evaluation in the usual
  way, cf., ~\cite{KleineBuningLettman99}.
  A literal is a Boolean variable~$x$ or its negation~$\neg
  x$.
  A \emph{CNF formula}~$\varphi$ is a set of \emph{clauses} interpreted as conjunction. A clause is a set of literals
  interpreted as disjunction. For a formula or clause~$X$, we
  abbreviate by $\var(X)$ the variables that occur~in~$X$.
An \emph{assignment} of~$\varphi$ is a mapping
$I: \var(\varphi) \rightarrow \{0,1\}$.
The formula~$\varphi(I)$ \emph{under assignment~$I$} is obtained
by removing every clause~$c$ from $\varphi$ that contains a literal set to~$1$
by $I$, and removing from every remaining clause of~$\varphi$ all literals set
to~$0$ by~$I$. An assignment~$I$ is \emph{satisfying} if
$\varphi(I)=\emptyset$.
\emph{Problem \cSAT} asks to output the number of satisfying assignments
of a formula.

We also allow \emph{equality formulas}, which are Boolean formulas, where variables are expressions using equality.
In more detail: Let $d$ be a fixed constant over domain~$\dom(v)$, where we call $d$ \emph{term constant}. 
Let $v$ and $v'$ be variables over some domain~$\dom(v)$ and $\dom(v')$, where we call $v$ and $v'$ \emph{term variables}.
Then, an \emph{equality formula}~$\beta$ is an expression of the form~$v{=}d$ or $v{=}v'$.
A \emph{term assignment~$J$} of equality formula~$\beta$ over term variables~$\tvar(\beta)$ assigns each domain variable~$v\in\tvar(\beta)$ a value over domain~$\dom(v)$.
The Boolean formula~$\beta(J)$ \emph{under term assignment~$J$} is
obtained as follows. First, we replace all expressions~$v{=}d$ in~$\beta$
by~$1$ if $J(v)=d$, all expressions~$v{=}v'$ by~$1$
if $J(v)=J(v')$, and by~$0$ otherwise. 
Second, we remove from the resulting clauses in~$\beta(J)$ each clause~$c$ that contains a literal set to~$1$.
Finally, we remove from every remaining clause in~$\beta(J)$ every literal that is set to~$0$.
We say a term assignment~$J$
is \emph{satisfying} if~$\beta(J)=\emptyset$.

\subsection{Tree Decomposition and Treewidth}\label{sec:prelimns:tds} %

  Treewidth is widely used for fine-grained complexity analyzes and to
  establish algorithms that provide tractability when bounding the
  treewidth. While it is only defined for graphs and hence widely used
  in graph theory~\cite{BodlaenderKoster08}, one can define graph
  representations of input instances for a variety of problems.
  Dedicated techniques then allow to solve problems from many domains
  such as propositional satisfiability~\cite{SamerSzeider10b},
  artificial intelligence~\cite{GottlobSzeider07}, knowledge
  representation~\cite{GottlobPichlerWei06},
  argumentation~\cite{FichteHecherMeier19}, non-monotonic
  reasoning~\cite{FichteHecherSchindler18a}, abduction in
  Datalog~\cite{GottlobPichlerWei07}, and databases~\cite{Grohe07},
  probabilistic inference~\cite{Dechter99} (under the name bucket
  elimination) including constraint satisfaction, Fourier and Gaussian
  elimination for solving linear equalities and inequalities, and
  combinatorial optimization. While theoretical conditional lower
  bound results seem to discourage using algorithms that exploit
  bounded treewidth~\cite{PanVardi06,FichteHecherPfandler20}, dynamic
  programming along tree decompositions or related decompositions has
  recently been used to establish practical
  algorithms~\cite{FichteHecherZisser19a,DudekPhanVardi20}.
  An algorithm that exploits small treewidth takes a tree
  decomposition, which is an arrangement of a graph into a tree, and
  evaluates the problem in parts, via dynamic programming (DP) on the
  tree decomposition.
  Informally speaking, the tree decomposition provides an evaluation
  ordering, which one employs by a problem specific algorithm where
  the runtime of combinatorial hard part is bounded by a (potentially
  exponential) function of the treewidth~$w$.
  Then, the underlying idea of treewidth is that it provides a
  measure for the closeness of a potential evaluation ordering which
  is simply a tree.
  Below, we provide the formal definitions of the notions tree
  decomposition and treewidth.
\begin{figure}[t]%
\includegraphics[scale=1.2]{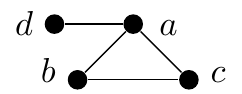}
\hspace{2em}%
\includegraphics[scale=1.2]{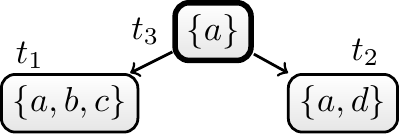}
\caption{Graph~$G$ %
  (left) with a TD~${\cal T}$ of graph~$G$
  (right).}%
\label{fig:graph-td}%
\end{figure}

A \emph{tree decomposition (TD)}~\cite{Kloks94a,CyganEtAl15} of a given graph~$G$ is a pair
$\TTT=(T,\chi)$ where $T$ is a rooted tree and $\chi$ is a mapping
which assigns to each node $t\in V(T)$ a set~$\chi(t)\subseteq V(G)$,
called \emph{bag}, such that (i) $V(G)=\bigcup_{t\in V(T)}\chi(t)$
and
$E(G)\subseteq\SB \{u,v\} \SM t\in V(T), \{u,v\}\subseteq \chi(t)\SE$;
and (ii) for each $r, s, t\in V(T)$, such that $s$ lies on the path
from~$r$ to $t$, we have $\chi(r) \cap \chi(t) \subseteq \chi(s)$. We
let $\width(\TTT) \eqdef \max_{t\in V(T)}\Card{\chi(t)}-1$.
The
\emph{treewidth} $\tw(G)$ of $G$ is the minimum $\width({\TTT})$ over
all TDs~$\TTT$ of $G$.
  For $k\in \mathcal{N}$ we can compute a tree decomposition of width~$k$ or
  output that none exists in
  time~$2^{\mathcal{O}{k^3}} \cdot \Card{V}$~\cite{BodlaenderKoster08}.
\begin{example}
Figure~\ref{fig:graph-td} depicts a graph~$G$ (left) 
and a TD~$\TTT$ of~$G$ (right) of width~$2$.
  The treewidth of~$G$ is also~$2$ since~$G$, contains
	a complete graph with~$3$ vertices~\cite{Kloks94a}.
\end{example}
  Next, we give the definition of a standard restriction for TDs, so
  called nice TDs. A nice TD ensures that one needs to consider only a
  small number of case distinctions in a DP algorithm later.
For a node~$t \in V(T)$, we say that $\type(t)$ is $\leaf$ if $t$ has
no children and~$\chi(t)=\emptyset$; $\join$ if $t$ has children~$t'$ and $t''$ with
$t'\neq t''$ and $\chi(t) = \chi(t') = \chi(t'')$; $\intr$
(``introduce'') if $t$ has a single child~$t'$,
$\chi(t') \subseteq \chi(t)$ and $|\chi(t)| = |\chi(t')| + 1$; $\rem$
(``removal'') if $t$ has a single child~$t'$,
$\chi(t') \supseteq \chi(t)$ and $|\chi(t')| = |\chi(t)| + 1$. If for
every node $t\in V(T)$, %
$\type(t) \in \{ \leaf, \join, \intr, \rem\}$, then the TD is called \emph{nice}.
  The conditions allow us to focus on each of the cases of our
  algorithms individually.
\section{Dynamic Programming on Tree Decompositions}\label{sec:dp}

  In the preliminaries, we gave definitions for tree decompositions
  and treewidth and stated a variety of application areas. We
  mentioned that treewidth is widely used as a structural measure to
  establish tractability results when we consider in addition to the
  input size of the instance also the treewidth.
  If we want to exploit small treewidth of an input instance by an
  algorithm in practice, one can design a so called-dynamic
  programming (DP) algorithm, which works as follows:
  Sub-problems are evaluated along the tree decomposition. At
  each node of the tree, information is gathered in tables.  A table
  contains tuples of a fixed form that are designed to ensure certain
  properties.  Then, a table algorithm maintains these tables during a
  post-order traversal. Thereby, it handles different cases according
  to the node contents of the TD and it ensures that properties
  required to solve the problem in parts sustain.
  The size of a table depends on the number of items in the bag, but
  is allowed to be exponential in the size of a bag.  Hence, the
  overall technique works in linear time in the size of the problem and
  exponential in the bag size.
  Intuitively, the tree decomposition fixes an order in which we
  evaluate our problem. 
  As a result, evaluating a problem along a tree decomposition allows
  for solving the problem at interest in parts, where the tree
  decomposition provides these parts and directs how solutions to the
  parts are supposed to be merged.

  More formally,
a solver based on \emph{dynamic programming (DP)} %
evaluates the input~$\mathcal{I}$ in parts along a given TD of a graph representation~$G$
of the input.
Thereby, for each node~$t$ of the TD, intermediate results are %
stored in a \emph{table}~$\tab{t}$. %
This is achieved by running a so-called \emph{table algorithm}~$\algo{A}$,
which is designed for a certain graph representation, 
and stores in~$\tab{t}$ results of problem parts of~$\mathcal{I}$,
thereby considering tables~$\tab{t'}$ for child nodes~$t'$ of~$t$. %
DP works for many problems~$\mathcal{P}$ as follows. %
\begin{enumerate}%
\item Construct a graph representation~$G$ of the given input instance~$\mathcal{I}$.
\item Heuristically compute a tree decomposition~$\TTT=(T,\chi)$ of~$G$.
\item\label{step:dp} Traverse the nodes in~$V(T)$ in
  post-order, i.e., perform a bottom-up traversal of~$T$.
  At every node~$t$ during post-order traversal, execute a table algorithm~$\algo{A}$ 
  that takes as input $t$, bag $\chi(t)$, a \emph{local instance}~$\mathcal{P}(t,\mathcal{I})=\mathcal{I}_t$ depending on~$\mathcal{P}$, as well as previously computed child tables of~$t$ and stores the result in~$\tab{t}$.
\item Interpret table~$\tab{n}$ for the root~$n$ of~$T$ in order to output the solution of~$\mathcal{I}$.
\end{enumerate}

  When specifying a DP algorithm for a specific problem such as \#SAT,
  it is often sufficient to provide the data structures and the table
  algorithm for the specific problem as the general outline of the DP
  works the same for most problems.
  Hence, we focus on table algorithms and their description in the
  following.
  Next, we state the graph representation and table algorithm that we
  need to solve the problem~$\mathcal{P}=\cSAT$~\cite{SamerSzeider10b}.
First, we  need the following graph representation.
The \emph{primal graph}~$G_\varphi$ of a formula~$\varphi$
has as vertices its variables, where two variables are joined by an edge
if they occur together in a clause of~$\varphi$.
Given formula~$\varphi$, a TD~$\mathcal{T}=(T,\chi)$ of~$G_\varphi$
and a node~$t$ of $T$.
Sometimes, we refer to the treewidth of the primal graph of a given formula
by the \emph{treewidth of the formula}.
Then, we
let local instance~$\cSAT(t, \varphi)=\varphi_t$ be $\varphi_t \eqdef \SB c \SM c \in \varphi, \var(c) \subseteq \chi(t)\SE$, which are the clauses entirely covered by~$\chi(t)$.

Table algorithm~$\algo{Sat}$ as presented in \algorithmcfname~\ref{alg:prim} shows all the cases that are needed to solve~\cSAT by means of DP over nice TDs.
Each table~$\tab{t}$ consist of rows of the form~$\langle I, c\rangle$,
where~$I$ is an assignment of~$\varphi_t$ and~$c$ is a counter. %
Nodes~$t$ with~$\type(t)=\leaf$ consist of the empty assignment and counter~$1$, cf., Line~1.
For a node~$t$ with introduced variable~$a\in\chi(t)$, we guess in Line~3 for each assignment~$\beta$ of the child table, whether~$a$ is set to true or to false, and ensure that~$\varphi_t$ is satisfied.
When an atom~$a$ is removed in node~$t$, we project assignments of child tables to~$\chi(t)$, cf., Line~5, and sum up counters of the same assignments.
For join nodes, counters of common assignments are multiplied, cf., Line~7.
  In Example~\ref{ex:running0} below, we explain the algorithm for a selected formula.

\begin{algorithm}[t]
  \KwData{Node~$t$, bag $\chi(t)$, clauses~$\varphi_t$, and a
    sequence $\langle \tab{1},\ldots \tab{\ell}\rangle$ of child tables.}
\KwResult{Table~$\tab{t}.$} \lIf(\hspace{-1em})
  {$\type(t) = \leaf$}{%
    $\tab{t} \eqdef \{ \langle
    \tuplecolor{\specialPredColor}{\emptyset},
    \tuplecolor{\statePredColor}{1} \rangle \}$%
  }%
  \uElseIf{$\type(t) = \intr$, and
    $a\hspace{-0.1em}\in\hspace{-0.1em}\chi(t)$ is introduced}{ %
    \makebox[3.3cm][l]{$\tab{t} \eqdef \{ \langle
      \tuplecolor{\specialPredColor}{I \cup \{a \mapsto 0\}},
      \tuplecolor{\statePredColor}{c} \rangle$
    }%
    \hspace{9em}$|\;
    \langle \tuplecolor{\specialPredColor}{I},
    \tuplecolor{\statePredColor}{c} \rangle \in \tab{1},
    \varphi_t(\tuplecolor{\specialPredColor}{I \cup \{a \mapsto 0\}})=\emptyset\} \cup$  \\
    \makebox[0.9cm][l]{}$\langle \tuplecolor{\specialPredColor}{I \cup \{a \mapsto 1\}}, \tuplecolor{\statePredColor}{c} \rangle\hspace{10em}
    |\;%
    \langle \tuplecolor{\specialPredColor}{I},
    \tuplecolor{\statePredColor}{c} \rangle \in \tab{1},
    \varphi_t(\tuplecolor{\specialPredColor}{I \cup \{a \mapsto 1\}})=\emptyset\}\hspace{-5em}$\\
  }\vspace{-0.05em}%
  \uElseIf{$\type(t) = \rem$, and $a \not\in \chi(t)$ is removed}{%
    \makebox[5cm][l]{$\tab{t} \eqdef \{ \langle
      \tuplecolor{\specialPredColor}{I\setminus\{a \mapsto 0, a \mapsto 1\}}$,
      \tuplecolor{\statePredColor}{%
        $\sum_{%
          c \in \text{C}(I) %
        }
        c \rangle\}$
      }%
    }
    \hspace{3.4em}$| \langle \tuplecolor{\specialPredColor}{I},
    \tuplecolor{\statePredColor}{\cdot} \rangle \in \tab{1}
    \}$\;
    \tcc{$\text{C}(I)$ is the set that contains all counters for which assignments~$J$ are the same as $I$ after we remove a from the assignment~$I$}
    $\text{C}(I) \eqdef \{ c \SM \langle {J}, {c}\rangle \in \tab{1}, J\setminus\{a \mapsto 0, a \mapsto 1\} = I\setminus\{a \mapsto 0, a \mapsto 1\} \SE$
  } %
  \uElseIf{$\type(t) = \join$}{%
    \makebox[3.3cm][l]{$\tab{t} \eqdef \{ \langle
      \tuplecolor{\specialPredColor}{I},
      \tuplecolor{\statePredColor}{c_1 \cdot c_2}
      \rangle$}\hspace{9em}$|\;\langle \tuplecolor{\specialPredColor}{I},
    \tuplecolor{\statePredColor}{c_1} \rangle \in \tab{1}, \langle
    \tuplecolor{\specialPredColor}{I},
    \tuplecolor{\statePredColor}{c_2} \rangle \in \tab{2}
    \}\hspace{-5em}$
    \vspace{-0.25em}
  } %
  \Return $\tab{t}$ \vspace{-0.25em}
  \caption{Table algorithm~$\algo{Sat}(t,\chi(t),\varphi_t,\langle \tab{1}, \ldots, \tab{\ell}\rangle)$ for solving \cSAT~\protect\cite{SamerSzeider10b}.}
  \label{alg:prim}
\end{algorithm}%

\begin{figure}[t]
\resizebox{1\textwidth}{!}{%
  \includegraphics{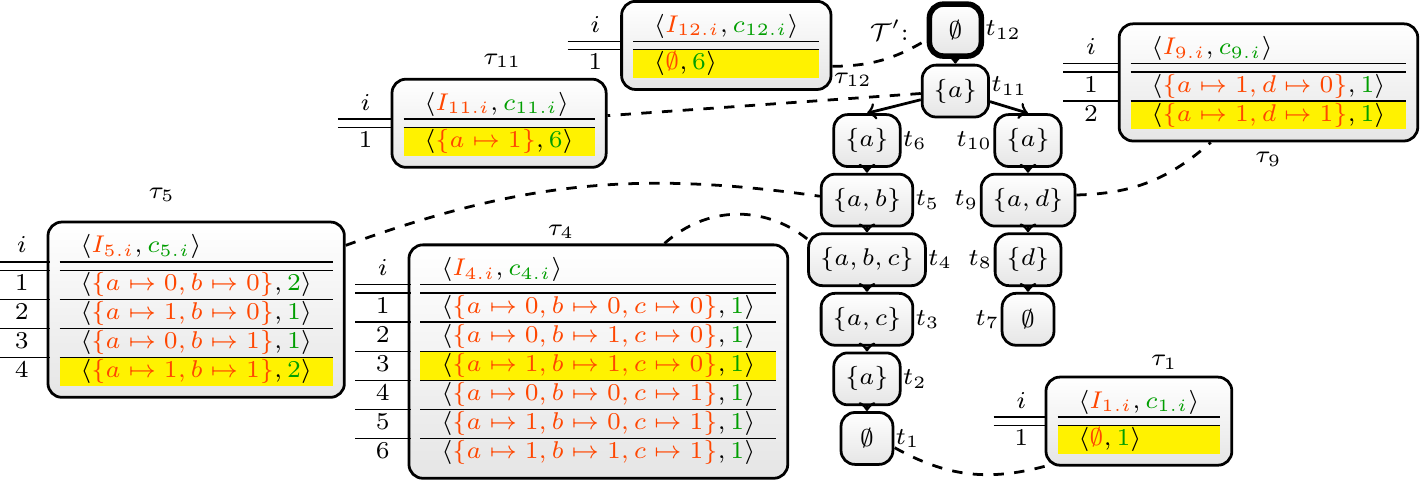}
}
\caption{Selected tables obtained by DP on~${\cal T}'$ for~$\varphi$ of Example~\ref{ex:running0} using algorithm~\algo{Sat}.}
\label{fig:running1}
\end{figure}

\medskip
\begin{example}\label{ex:running0}\vspace{-1.25em}%
Consider
  formula~$\varphi\eqdef \{\overbrace{\{\neg a, b, c\}}^{c_1},
  \overbrace{\{a, \neg b, \neg c\}}^{c_2}, \overbrace{\{a,
    d\}}^{c_3}, \overbrace{\{a, \neg d\}}^{c_4}\}$.
  Satisfying assignments of formula~$\varphi$ are, e.g., 
  $\{a\mapsto 1,b\mapsto 1, c\mapsto 0, d\mapsto 0\}$, $\{a\mapsto 1, b\mapsto 0,c\mapsto 1, d\mapsto 0\}$ or $\{a\mapsto 1, b\mapsto 1,c\mapsto 1, d\mapsto 1\}$.
  In total, there are 6 satisfying assignments of~$\varphi$. 
  Observe that graph~$G$ of Figure~\ref{fig:graph-td} depicts
  the primal graph~$G_\varphi$ of~$\varphi$.
  Intuitively, ${\cal T}$ of Figure~\ref{fig:graph-td} allows to
  evaluate formula~$\varphi$ in parts. 
  Figure~\ref{fig:running1} illustrates a nice TD~$\TTT'=(\cdot, \chi)$ of the 
  primal graph~$G_\varphi$ and
  tables~$\tab{1}$, $\ldots$, $\tab{12}$ that are obtained during the
  execution of~${\algo{Sat}}$ on nodes~$t_1,\ldots,t_{12}$.
  We assume that each row in a table $\tab{t}$ is identified by a
  number,~i.e., row $i$ corresponds to
  $\vec{u_{t.i}} = \langle I_{t.i}, c_{t.i} \rangle$.

  Table~$\tab{1}=\SB \langle\emptyset, 1\rangle \SE$ has
  $\type(t_1) = \leaf$.
  Since $\type(t_2) = \intr$, we construct table~$\tab{2}$
  from~$\tab{1}$ by taking~$I_{1.i}\cup\{a\mapsto 0\}$ and $I_{1.i}\cup \{a \mapsto 1\}$ for
  each~$\langle I_{1.i}, c_{1.i}\rangle \in \tab{1}$. Then,
  $t_3$ introduces $c$ and $t_4$ introduces $b$.
  $\varphi_{t_1}=\varphi_{t_2}=\varphi_{t_3} = \emptyset$, but since
  $\chi(t_4) \subseteq \var(c_1)$ we have
  $\varphi_{t_4} = \{c_1,c_2\}$ for $t_4$.
  In consequence, for each~$I_{4.i}$ of table~$\tab{4}$, we have
  $\{c_1,c_2\}({{I_{4.i}}})=\emptyset$ since \algo{Sat} enforces
  satisfiability of $\varphi_t$ in node~$t$.  
  Since $\type(t_5) = \rem$, we remove variable~$c$ from all
  elements in $\tab{4}$ and sum up counters accordingly to construct $\tab{5}$. 
  Note that we have
  already seen all rules where $c$ occurs and hence $c$ can no
  longer affect interpretations during the remaining traversal. We
  similarly create $\tab{6}=\{\langle \{a\mapsto 0\}, 3 \rangle, \langle \{a \mapsto 1\}, 3 \rangle\}$
  and~$\tab{{10}}=\{\langle \{a \mapsto 1\}, 2 \rangle\}$.
  Since $\type(t_{11})=\join$, we build table~$\tab{11}$ by taking
  the intersection of $\tab{6}$ and $\tab{{10}}$. Intuitively, this
  combines assignments agreeing on~$a$, where counters are multiplied accordingly.
  By definition (primal graph and TDs), for every~$c \in \varphi$,
  variables~$\var(c)$ occur together in at least one common bag.
  Hence, %
  since
  $\tab{12} = \{\langle \emptyset, 6 \rangle \}$, we can reconstruct for example
  model~$\{a\mapsto 1,b \mapsto 1, c\mapsto 0, d\mapsto 1\} = I_{11.1} \cup I_{5.4} \cup I_{9.2}$ of~$\varphi$ using highlighted (yellow) rows in Figure~\ref{fig:running1}.
  On the other hand, if~$\varphi$ was unsatisfiable, $\tab{12}$ would contain no values, i.e., $\tab{12}=\emptyset$. %
\end{example}%

\section{Dynamic Programming on Tree Decompositions Expressed in Relational Algebra}

  While algorithms that run dynamic programming on bounded treewidth
  can be quite useful for efficient problem solving in practice,
  implementations turn out to be tedious already for problems such as
  the propositional satisfiability problem.
  In the following of the paper, we aim for rapid prototyping with
  dynamic programming by a declarative approach that ideally uses
  existing systems, gets parallel execution for free, and remains
  fairly efficient.

  In the previous section, we explained that the traversal of the tree
  decomposition and the overall methodology of the procedure stays the
  same.  But the core of dynamic programming on tree decompositions for
  various problems is mostly the specification of the table algorithm
  that modifies a table based on previously computed tables.  
  Hence, one can often focus on the table algorithms and their
  descriptions.
  When recalling basics from databases 101 from undergraduate
  studies~\cite{ElmasriNavathe16} and taking a closer look on
  Algorithm~\ref{alg:prim} above, we can immediately spot that we are
  effectively describing a query on existing data that produces a new
  table by Algorithm~\ref{alg:prim}.
  This motivates our idea to use a database management system to
  execute the query and specify the query in SQL.
  Before we can proceed with our idea to use databases for executing
  DP algorithms, we take a step back and recall that the theory of SQL
  queries is based on relational algebra.

  Relational algebra allows us to describe our algorithms and later
  use SQL encodings for specifying the table algorithm.
  The intermediate step of stating the algorithm in a relation algebra
  description is twofold. First, we can immediately see the connection
  between the algorithms given in the literature, which allows us to
  use the existing algorithms without reproving all properties.
  Second, we obtain a compact mathematical description, which is not
  just a lengthy and technical SQL query that might be hard to
  understand to researchers from the community who are usually not
  very familiar with practical databases and the usage of query
  languages.

\subsection{Relational Algebra}%
  Before we start with details on our approach, we briefly recall
  basics in relational algebra.
  The classical relational algebra was introduced by
  Codd~\shortcite{Codd70} as a mathematical framework for manipulating
  relations (tables). Since then, relational algebra serves as the
  formal background and theoretical basis in relational databases and
  their standard language \emph{SQL (Structured Query Language)} for
  querying tables~\cite{Ullman89}.
  In fact, in the following, we need extended constructs, which have
  not been defined in the original framework by Codd, but are standard
  notions in databases nowadays~\cite{ElmasriNavathe16}.
  For the understanding later, we would like to mention that the SQL
  table model and relational algebra model slightly differ. The SQL
  table model is a bag (multiset) model, rather than a
  set~\cite[Chapter 5]{Garcia-MolinaUllmanWidom09}.
  Below we also use extended projection and aggregation by grouping.
  Sometimes these are defined on bags. We avoid this in the
  definitions in order to keep the algorithms close to the formal set
  based notation.
  Finally, we would like to emphasize that we are not using relation
  algebra here as developed by Alfred Tarski for the field of abstract
  algebra, but really relational algebra as used in database
  applications and theory.

A \emph{column}~$a$ is of a certain finite \emph{domain~$\dom(a)$}.
Then, a \emph{row}~$r$ over set~$\attr(r)$ of columns
is a set of pairs of the form~$(a, v)$ with~$a\in\attr(r),v\in \dom(a)$ such that for each~$a\in \attr(r)$, there is exactly one~$v\in\dom(a)$ with~$(a,v)\in r$.
In order to \emph{access} the value~$v$ of an attribute~$a$ in a row~$r$,
we sometimes write~$r.a$, which returns the unique value~$v$ with~$(a,v)\in r$.
A \emph{table~$\tab{}$} is a finite set of rows~$r$ over set~$\attr(\tab{})\eqdef\attr(r)$ of columns, using domain~$\dom(\tab{})\eqdef \bigcup_{a\in \attr(\tab{})}\dom(a)$.
We define \emph{renaming} of~$\tab{}$, given a set~$A$ of columns and a bijective mapping~$m:\attr(\tab{}) \rightarrow A$ with $\dom(a)=\dom(m(a))$ for~$a\in\attr(\tab{})$, by~$\rho_m(\tab{}) \eqdef \{(m(a),v) \mid (a,v)\in \tab{}\}$. In SQL, renaming can be achieved by means of the {\ttfamily AS} keyword.

\emph{Selection} of rows in $\tab{}$ according to a given equality formula~$\varphi$ 
over term variables~$\attr(\tab{})$
is defined\footnote{We abbreviate for binary $v\in\attr(\tab{})$ with~$\dom(v)=\{0,1\}$, $v{=}1$ by~$v$ and~$v{=}0$ by~$\neg v$.} by~$\sigma_{\varphi}(\tab{})\eqdef \{ r \mid r\in \tab{}, \varphi(\ass(r))=\emptyset \}$,
where function~$\ass$ provides the \emph{corresponding term assignment} of a given row~$r\in\tab{}$.
Selection in SQL is specified using keyword {\ttfamily WHERE}.
Given a relation~$\tab{}'$ with~$\attr(\tab{}')\cap\attr(\tab{})=\emptyset$. Then, we refer to the \emph{cross-join} by~$\tab{}\times \tab{}'\eqdef \{ r\cup r' \mid r\in \tab{}, r'\in \tab{}'\}$.
Further, a \emph{$\theta$-join} (according to $\varphi$) corresponds to~$\tab{} \bowtie_\varphi \tab{}' \eqdef \sigma_\varphi(\tab{}\times \tab{}')$. Interestingly, in SQL a $\theta$-join can be achieved by specifying the two tables (cross-join) and adding the selection according to~$\varphi$ by means of {\ttfamily WHERE}.

Assume in the following a set~$A\subseteq \attr(\tab{})$ of columns.
Then, we let table~$\tab{}$ \emph{projected to~$A$} be given by $\Pi_{A}(\tab{})\eqdef \{r_A \mid r\in \tab{}\}$, where~$r_A \eqdef \{(a, v) \mid (a, v) \in r, a \in A\}$.
This concept of projection can be lifted to \emph{extended projection~$\dot\Pi_{A,S}$}, where we assume in addition to~$A$, a set~$S$ of expressions of the form~$a \leftarrow f$, such that $a\in \attr(\tab{})\setminus A$, $f$ is an arithmetic function that takes a row~$r\in \tab{}$, and there is at most one such expression for each $a\in \attr(\tab{})\setminus A$ in~$S$.
Formally, we define $\dot\Pi_{A,S}(\tab{})\eqdef \{r_A \cup r^S \mid r\in \tab{}\}$ with~$r^S \eqdef \{(a, f(r)) \mid a \in \attr(r), (a \leftarrow f) \in S\}$.
SQL allows to specify (extended) projection directly after initiating an SQL query with the keyword {\ttfamily SELECT}.

Later, we use \emph{aggregation by grouping~$_A G_{(a\leftarrow g)}$}, where we assume~$a\in\attr(\tab{})\setminus A$ and a so-called \emph{aggregate function~$g: 2^\tab{}\rightarrow \dom(a)$}, which intuitively takes a table of (grouped) rows. Therefore, we let~$_A G_{(a\leftarrow g)}(\tab{})\eqdef \{r\cup \{(a,g(\tab{}[r]))\} \mid r\in\Pi_{A}(\tab{})\}$, where $\tab{}[r]\eqdef\{r'\mid r'\in \tab{}, r\subseteq r'\}$.
For this purpose, we use for a set~$S\subseteq \mathbb{S}$ of integers, 
the functions~$\text{\ttfamily SUM}$ for summing up values in~$S$, $\text{\ttfamily MIN}$ for providing the smallest integer in~$S$, as well as $\text{\ttfamily MAX}$ for obtaining the largest integer in~$S$, which are often used for aggregation in this context.
The SQL standard uses projection ({\ttfamily SELECT}) to specify~$A$ as well as the aggregate function~$g$, such that these two parts are distinguished by means of the keyword {\ttfamily GROUP BY}.

\begin{example}
Assume a table~$\tab{1}\eqdef\{r_1, r_2, r_3\}$ of 2 columns~$a,b$ over Boolean domain $\dom(a)=\dom(b)=\{0,1\}$, where~$r_1\eqdef\{(a,1), (b,1)\}$,
$r_2\eqdef\{(a,0), (b,0)\}$,
$r_3\eqdef\{(a,0), (b,1)\}$.
\begin{figure*}[h!]
\centering\includegraphics{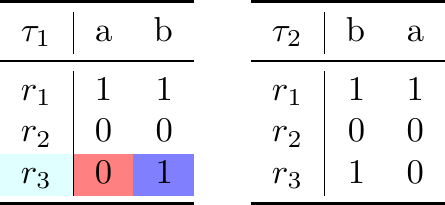}
\end{figure*}

\noindent Then, $r_3.a= {\color{red!50}0}$ and~$r_3.b={\color{blue!50}1}$. 
Rows can be swapped by renaming and we let~$\tab{2}\eqdef \rho_{\{a\mapsto b, b\mapsto a\}}\tab{1}$.

\noindent Observe that, e.g., $\rho_{\{a\mapsto b, b\mapsto a\}}(\{r_3\})$ corresponds to $\{\{(a,1),(b,0)\}\}$,~i.e., considering $r_3$ 
and swapping $a$ and $b$.
We select rows by using the selection~$\sigma$. For example, if we
want to select rows where~$b=1$ (colored in blue) we can
use~$\sigma_{b=1}(\tab{1})$.
\begin{figure*}[h!]
\centering\includegraphics{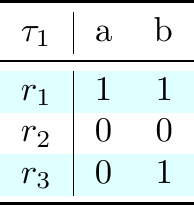}
\end{figure*}

\noindent Hence, applying $\sigma_{b=1}(\tab{1})$ results in~$\{r_1,r_3\}$.
Table~$\tab{1}$ can be~$\theta$-joined with~$\tab{2}$, but before, we need to have disjoint columns, 
which we obtain by renaming each column~$c$ to a fresh column~$c'$ as below by $\rho_{a\mapsto a', b\mapsto b'}\tab{2}$.
Then, $\tau_3 \eqdef \tab{1}\bowtie_{a=a'\wedge b=b'}(\rho_{a\mapsto a', b\mapsto b'}\tab{2})$.
\begin{figure*}[h!]
\centering\includegraphics{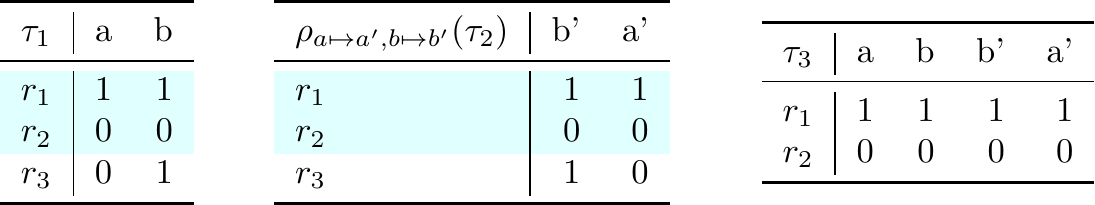}
\end{figure*}

\noindent Consequently, we have~$\tau_3 =\{\{(a,0),(a',0),(b,0),(b',0)\}, \{(a,1),(a',1),(b,1),(b',1)\}\}$.
Extended projection allows not only to filter certain columns,
but also to add additional columns.
As a result, if we only select column~$a$ of each row of~$\tab{1}$, but add a fresh column~$c$ holding the sum of the values for~$a$ and~$b$, then
$\dot\Pi_{\{a\},\{c\leftarrow a+b\}}\tab{1}$ corresponds to~$\{\{(a,1), (c,2)\}, \{(a,0),(c,0)\},$ $\{(a,0),(c,1)\}\}$.
\begin{figure*}[h!]
\centering\includegraphics{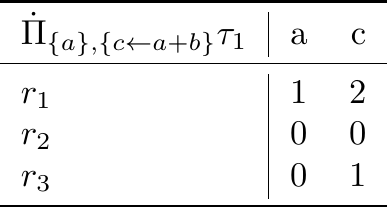}
\end{figure*}

\noindent Grouping~$\tab{1}$ according to the value of column~$a$, where we aggregate each group by summing up the values of columns~$b$ in a fresh column~$d$, results in~$_{\{a\}}G_{d\,\leftarrow\, \tab{}\mapsto\text{\ttfamily SUM}(\{r.b\mid r\in\tab{}\})}(\tau_1)$, which simplifies to~$\{\{(a,1), (d,1)\}, \{(a,0),\{d,1)\}\}$ as illustrated below.
\begin{figure*}[h!]
\centering\includegraphics{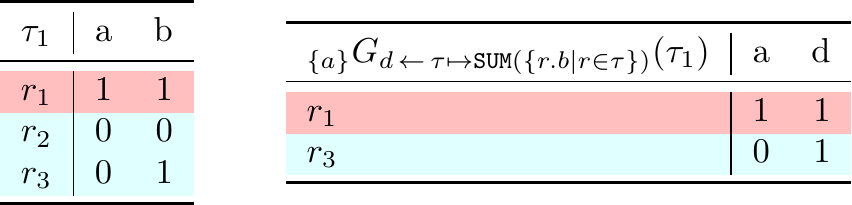}
\end{figure*}

\end{example}

\begin{algorithm}[b]
  \KwData{Node~$t$, bag $\chi(t)$, local formula~$\varphi_t$,
    sequence $\langle \tab{1},\ldots \tab{\ell}\rangle$ of child tables.}
\KwResult{Table $\tab{t}.$} \lIf(\hspace{-1em})
  {$\type(t) = \leaf$}{%
    $\tab{t} \eqdef \tuplecolor{\inputPredColor}{\{ \{{(\text{cnt}, 1)}\} \}}$ \label{line:leaf}%
  }%
  \uElseIf{$\type(t) = \intr$, and
    $a\hspace{-0.1em}\in\hspace{-0.1em}\chi(t)$ is introduced}{ %
    $\tab{t}\eqdef \tab{1} \bowtie_{\tuplecolor{\inputPredColor}{\varphi_t}} \tuplecolor{\inputPredColor}{\{\{(\cid{a},0)\},\{(\cid{a},1)\}\}}$\label{line:intr}
    \hspace{-5em}\vspace{-0.05em}
  }\vspace{-0.05em}%
  \uElseIf{$\type(t) = \rem$, and $a \not\in \chi(t)$ is removed}{%
    $\tab{t} \eqdef {_{\chi(t)}G}_{\tuplecolor{\inputPredColor}{\text{cnt} \,\leftarrow\, \tab{}\mapsto \text{\ttfamily SUM}(\{r.\text{cnt}\mid r\in\tab{}\})}}(\Pi_{{\attr(\tab{1})\setminus\{\cid{a}}\}}\tab{1})$\label{line:rem} \hspace{-5em} \vspace{-0.1em}
  } %
  \uElseIf{$\type(t) = \join$}{%
    \makebox[3.3cm][l]{$\tab{t} \eqdef \dot\Pi_{\chi(t),\tuplecolor{\inputPredColor}{\{\text{cnt} \leftarrow\text{cnt} \cdot \text{cnt}'\}}}(\tab{1} \bowtie_{\tuplecolor{\inputPredColor}{}{\bigwedge_{u\in\chi(t)}\cid{u}= \cid{u}'}} \rho_{\hspace{-.75em}\bigcup\limits_{u\in \attr(\tab{2})}\hspace{-.75em}\{\cid{u}\mapsto \cid{u}'\}}\tab{2})$} \label{line:join} %
    \hspace{-5em}%
    \vspace{-0.75em}
  } %
  \Return $\tab{t}$ \vspace{-0.25em}
  \caption{Table algorithm~$\algo{Sat'}(t,\chi(t),\varphi_t,\langle \tab{1}, \ldots, \tab{\ell}\rangle)$ for solving \cSAT.}
  \label{alg:primdb}
\end{algorithm}%

\subsection{Table Algorithms in Relational Algebra}
Now, we are in the position to use relational algebra instead of set
theory based notions to describe how tables are obtained during
dynamic programming.
The step from set notation to relational algebra is driven by the
observation that in these table algorithms one can identify recurring
patterns and one mainly has to adjust problem-specific parts of it.
We continue the description with our problem~\cSAT.  
We picture tables~$\tab{t}$ for each TD node~$t$ as relations, where~$\tab{t}$ distinguishes a unique column~$\cid{x}$ for each~$x\in\chi(t)$.
In addition, we require a column \emph{cnt} for counting in~\cSAT, or
a column for modeling costs or weights in case of optimization
problems.

\algorithmcfname~\ref{alg:primdb} presents a table algorithm for problem~\cSAT that is equivalent to \algorithmcfname~\ref{alg:prim}, but relies on relational algebra for computing tables.
  Since our description in relation algebra yields the same results as
  the set based-notation above, we omit formal correctness
  proofs. Nonetheless, we briefly explain below why both notations are
  identical.
 We highlight the crucial parts by coloring \algorithmcfname~\ref{alg:prim}.
In particular, one typically derives for nodes~$t$ with~$\type(t)=\leaf$, a fresh initial table $\tab{t}$, cf., Line~\ref{line:leaf} of \algorithmcfname~\ref{alg:primdb}.
Then, whenever a variable~$a$ is introduced, such algorithms often use~$\theta$-joins 
with a fresh initial table for the introduced variable~$a$. Hence, the new column represents the potential values for variable~$a$. In Line~\ref{line:intr}, the selection of the $\theta$-join is performed according to~$\varphi_t$,~i.e., corresponding to the local instance of~\cSAT.
Further, for nodes~$t$ with~$\type(t)=\rem$, these table algorithms typically need projection. 
In case of \algorithmcfname~\ref{alg:primdb}, Line~\ref{line:rem} also needs grouping to sum up the counters for those rows of~$\tab{1}$ that concur in~$\tab{t}$.
Thereby, rows are grouped according to values of columns~$\chi(t)$ and we keep only one  row per group in table~$\tab{}$, where the fresh counter \emph{cnt} is the sum among all counters in~$\tab{}$.
Finally, in Line~\ref{line:join} for a node~$t$ with~$\type(t)=\join$, we use extended projection and $\theta$-joins, where we join on the same truth assignments. This  allows us later to leverage database technology for a usually expensive operation. %
Extended projection is needed for multiplying the counters of the two rows containing the same assignment.

\subsection{Table algorithms for selected problems}

  Dynamic programming algorithms are known for a variety of problems. 
  Standard texts in the area of parameterized algorithms and
  introductory lectures provide various specifications. For formal
  properties and detailed algorithm descriptions, we refer to other
  works~\cite{Bodlaender88}, \cite[Chapter 7]{CyganEtAl15},
  \cite{Dechter99}, \cite{BannachBerndt19}.
  Below, we present the table algorithms for a selection of
  combinatorial problems in relational algebra notation.
  In order to simplify the presentation, we assume that the instance
  is given by~$\mathcal{I}$ and that the used tree decomposition is
  nice and given by $\mathcal{T}=(T,\chi)$. If the problem is a
  graph problem $\mathcal{T}$ is a TD of~$\mathcal{I}$, otherwise we implicitly assume that $\mathcal{T}$ is a TD of the primal graph of instance~$\mathcal{I}$.
  For graph problems~$\mathcal{I}$ and a given node~$t$ of~$T$, we refer to the local instance
  of~$\mathcal{I}=G=(V,E)$ by \emph{local graph~$G_t$} and define it
  by $G_t \eqdef (V\cap\chi(t), E\cap [\chi(t)\times\chi(t)])$.

\subsubsection*{Problem~$\cTCOL$}
\begin{algorithm}[ht]
  \KwData{Node~$t$, bag $\chi(t)$, local graph~$G_t$,
    and a sequence $\langle \tab{1},\ldots \tab{\ell}\rangle$ of child tables.}
\KwResult{Table $\tab{t}.\hspace{-5em}$} \lIf(\hspace{-1em})
  {$\type(t) = \leaf$}{%
    $\tab{t} \eqdef \tuplecolor{\inputPredColor}{\{(\text{cnt}, 1)\}}$ \label{line:col:leaf2}%
  }%
  \uElseIf{$\type(t) = \intr$, and
    $a\hspace{-0.1em}\in\hspace{-0.1em}\chi(t)$ is introduced}{ %
    $\tab{t}\eqdef %
\tab{1} \bowtie_{\tuplecolor{\inputPredColor}{\bigwedge_{\{u,v\}\in E(G_t)} u\neq v}} \tuplecolor{\inputPredColor}{\{\{(a, 0)\}, \{(a, 1)\}, \ldots, \{(a,o)\}\}}$ \label{line:col:intr2}
    \hspace{-5em}\vspace{-0.05em}
  }\vspace{-0.05em}%
  \uElseIf{$\type(t) = \rem$, and $a \not\in \chi(t)$ is removed}{%
    $\tab{t} \eqdef {_{\chi(t)}G}_{\tuplecolor{\inputPredColor}{\text{cnt}\,\leftarrow\,\tab{}\mapsto\text{\ttfamily SUM}(\{r.\text{cnt}\mid r\in\tab{}\})}}(\Pi_{\attr(\tab{1})\setminus\{\cid{a}\}}\tab{1})$ \label{line:col:rem2}\hspace{-5em} \vspace{-0.1em}
  } %
  \uElseIf{$\type(t) = \join$}{%
    \makebox[3.3cm][l]{$\tab{t} \eqdef \dot\Pi_{\chi(t), \tuplecolor{\inputPredColor}{\{\text{cnt} \leftarrow\text{cnt} \cdot \text{cnt}'\}}}(\tab{1} \bowtie_{\tuplecolor{\inputPredColor}{}{\bigwedge_{u\in\chi(t)}\cid{u}= \cid{u}'}} \rho_{\hspace{-0.75em}\bigcup\limits_{u\in \attr(\tab{2})}\hspace{-0.75em}\{\cid{u}\mapsto \cid{u}'\}}\tab{2})$} \label{line:col:join2} %
    \hspace{-5em}%
    \vspace{-0.75em}
  }
  \Return $\tab{t}$ \vspace{-0.25em}
  \caption{Table algorithm~$\algo{Col}(t,\chi(t),G_t,\langle \tab{1}, \ldots, \tab{\ell}\rangle)$ for solving~$\cTCOL$.}
  \label{alg:col}
\end{algorithm}%

For a given graph instance~$\mathcal{I}=G=(V,E)$, an \emph{$o$-coloring} is a mapping~$\iota: V \rightarrow \{1,\ldots,o\}$
such that for each edge~$\{u,v\}\in E$, we have~$\iota(u)\neq \iota(v)$.
Then, the problem~$\cTCOL$ asks to count the number of $o$-colorings of~$G$,
whose local instance~$\cTCOL(t,G)$ is the local graph $G_t$.
The table algorithm for this problem~$\cTCOL$ is given in Algorithm~\ref{alg:col}.
Similarly to Algorithm~\ref{alg:primdb}, for (empty) leaf nodes, 
the counter~\emph{cnt} is set to~$1$ in Line~\ref{line:col:leaf2}.
Whenever a vertex~$a$ is introduced, in Line~\ref{line:col:intr2}, one of the $o$ many color values for~$a$ are guessed and $\theta$-joined with the table~$\tab{1}$ for the child node of~$t$ such that only colorings with different values for two \emph{adjacent} vertices are kept.
Similarly to Algorithm~\ref{alg:primdb}, whenever a vertex~$a$ is removed, Line~\ref{line:col:rem2} ensures that the column for~$a$ is removed
and that counters~\emph{cnt} are summed up for rows that concur due to the
removal of column~$a$.
Then, the case for join nodes in Line~\ref{line:col:join2} is again analogous to Algorithm~\ref{alg:primdb}, where only rows with the same colorings in both child tables
are kept and counters~\emph{cnt} are multiplied accordingly.

\subsubsection*{Problem~\VC}
\begin{algorithm}[ht]
  \KwData{Node~$t$, bag $\chi(t)$, local graph~$G_t$,
    and a sequence $\langle \tab{1},\ldots \tab{\ell}\rangle$ of child tables.}
\KwResult{Table $\tab{t}.\hspace{-5em}$} \lIf(\hspace{-1em})
  {$\type(t) = \leaf$}{%
    $\tab{t} \eqdef \tuplecolor{\inputPredColor}{\{(\text{card}, 0)\}}$ \label{line:vc:leaf2}%
  }%
  \uElseIf{$\type(t) = \intr$, and
    $a\hspace{-0.1em}\in\hspace{-0.1em}\chi(t)$ is introduced}{ %
    $\tab{t}\eqdef %
\tab{1} \bowtie_{\tuplecolor{\inputPredColor}{\bigwedge_{\{u,v\}\in E(G_t)} u \vee v}} \tuplecolor{\inputPredColor}{\{\{(a, 0)\}, \{(a, 1)\}\}}$ \label{line:vc:intr2}
    \hspace{-5em}\vspace{-0.05em}
  }\vspace{-0.05em}%
  \uElseIf{$\type(t) = \rem$, and $a \not\in \chi(t)$ is removed}{%
    $\tab{t} \eqdef {_{\chi(t)}G}_{\tuplecolor{\inputPredColor}{\text{card}\,\leftarrow\, \tab{}\mapsto\text{\ttfamily MIN}(\{r.\text{card} + r.a\mid r\in\tab{}\})}}(\Pi_{\attr(\tab{1})\setminus\{\cid{a}\}}\tab{1})$ \label{line:vc:rem2}\hspace{-5em} \vspace{-0.1em}
  } %
  \uElseIf{$\type(t) = \join$}{%
    \makebox[3.3cm][l]{$\tab{t} \eqdef \dot\Pi_{\chi(t),\tuplecolor{\inputPredColor}{\{\text{card} \leftarrow\text{card} + \text{card}'\}}}(\tab{1} \bowtie_{\tuplecolor{\inputPredColor}{}{\bigwedge_{u\in\chi(t)}\cid{u}= \cid{u}'}} \rho_{\hspace{-0.75em}\bigcup\limits_{u\in \attr(\tab{2})}\hspace{-0.75em}\{\cid{u}\mapsto \cid{u}'\}}\tab{2})$} \label{line:vc:join2} %
    \hspace{-5em}%
    \vspace{-0.75em}
  }
  \Return $\tab{t}$ \vspace{-0.25em}
  \caption{Table algorithm~$\algo{VC}(t,\chi(t),G_t,\langle \tab{1}, \ldots, \tab{\ell}\rangle)$ for solving~$\VC$.}
  \label{alg:vc}
\end{algorithm}%
Given a graph instance~$\mathcal{I}=G=(V,E)$, a \emph{vertex cover} is a set of vertices~$C\subseteq V$ of~$G$
such that for each edge~$\{u,v\}\in E$, we have~$\{u,v\}\cap C\neq \emptyset$.
Then, \VC asks to find the minimum cardinality~$|C|$ among all vertex covers~$C$, i.e., $C$ is such that there is no vertex cover~$C'$ with~$|C'| < |C|$. 
Local instance~$\VC(t,G)\eqdef G_t$, where the local graph~$G_t$ is defined above. 
We use an additional column \emph{card} for storing cardinalities.
The table algorithm for solving \VC is provided in Algorithm~\ref{alg:vc},
where, for leaf nodes the cardinality is $0$, cf., Line~\ref{line:vc:leaf2}.
Then, when introducing vertex~$a$, we guess in Line~\ref{line:vc:intr2} whether~$a$ shall be in the vertex cover or not,
and enforce that for each edge of the local instance at least one
of the two endpoint vertices has to be in the vertex cover.
Note that the additional cardinality column only takes removed vertices into account.
More precisely, when a vertex $a$ is removed, we group in Line~\ref{line:vc:rem2} according to the bag columns~$\chi(t)$, where the fresh cardinality value is the minimum
cardinality (plus 1 for~$a$ if~$a$ shall be in the vertex cover), 
among those rows that concur due to the removal of~$a$.
The join node is similar to before, but in Line~\ref{line:vc:join2} we additionally need to sum up
the cardinalities of two adjoining child table rows.

\subsubsection*{Problem~\MSAT}

\begin{algorithm}[ht]
  \KwData{Node~$t$, bag $\chi(t)$, local instance~$\mathcal{I}_t=(\varphi_t, \psi_t)$,
    and a sequence $\langle \tab{1},\ldots \tab{\ell}\rangle$ of child tables.}
\KwResult{Table $\tab{t}.\hspace{-5em}$} \lIf(\hspace{-1em})
  {$\type(t) = \leaf$}{%
    $\tab{t} \eqdef \tuplecolor{\inputPredColor}{\{(\text{card}, 0)\}}$ \label{line:msat:leaf2}%
  }%
  \uElseIf{$\type(t) = \intr$, and
    $a\hspace{-0.1em}\in\hspace{-0.1em}\chi(t)$ is introduced}{ %
    $\tab{t}\eqdef %
\tab{1} \bowtie_{\tuplecolor{\inputPredColor}{\varphi_t}} \tuplecolor{\inputPredColor}{\{\{(a, 0)\}, \{(a, 1)\}\}}$ \label{line:msat:intr2}
    \hspace{-5em}\vspace{-0.05em}
  }\vspace{-0.05em}%
  \uElseIf{$\type(t) = \rem$, $a \not\in \chi(t)$ is removed, and~$\psi'$ are removed local soft-clauses}{%
    $\tab{t} \eqdef {_{\chi(t)}G}_{\tuplecolor{\inputPredColor}{\text{card}\,\leftarrow\,\tab{}\mapsto\text{\ttfamily MAX}(\{r.\text{card} + \Sigma_{c\in\psi'\hspace{-.1em},\,c(\ass(r))=\emptyset} 1\mid r\in\tab{}\})}}(\Pi_{\attr(\tab{1})\setminus\{\cid{a}\}}\tab{1})$ \label{line:msat:rem2}\hspace{-5em} \vspace{-0.15em}
  } %
  \uElseIf{$\type(t) = \join$}{%
    \makebox[3.3cm][l]{$\tab{t} \eqdef \dot\Pi_{\chi(t),\tuplecolor{\inputPredColor}{\{\text{card}\leftarrow\text{card} + \text{card}'\}}}(\tab{1} \bowtie_{\tuplecolor{\inputPredColor}{}{\bigwedge_{u\in\chi(t)}\cid{u}= \cid{u}'}} \rho_{\hspace{-0.75em}\bigcup\limits_{u\in \attr(\tab{2})}\hspace{-0.75em}\{\cid{u}\mapsto \cid{u}'\}}\tab{2})$} \label{line:msat:join2} %
    \hspace{-5em}%
    \vspace{-0.75em}
  }
  \Return $\tab{t}$ \vspace{-0.25em}
  \caption{Table algorithm~$\algo{MSat}(t,\chi(t),\mathcal{I}_t,\langle \tab{1}, \ldots, \tab{\ell}\rangle)$ for solving~$\MSAT$.}
  \label{alg:msat}
\end{algorithm}%
Given Boolean formulas~$\varphi$ and $\psi$, an instance of problem~$\MSAT$ is of the form~$\mathcal{I}=(\varphi, \psi)$ and we assume that~$\mathcal{T}$ is a TD of primal graph~$G_{\varphi\cup\psi}$.
A solution to~\MSAT is a satisfying assignment~$I$ of \emph{hard-clauses}~$\varphi$ such that~$\Card{\SB c\SM c\in \psi, c(I)=\emptyset \SE}$ is maximized, i.e., $I$ is a satisfying assignment of~$\varphi$ that satisfies the maximum number of \emph{soft-clauses}~$\psi$ among all satisfying assignments of~$\varphi$. %
We define the local instance~$\mathcal{I}_t\eqdef(\varphi_t, \psi_t)$ consisting of local formula~$\varphi_t$, referred to by~$\emph{local hard-clauses}$ and local formula~$\psi_t$, called~$\emph{local soft-clauses}$.

The table algorithm for problem~\MSAT is given in Algorithm~\ref{alg:msat},
where we use column~\emph{card} for holding satisfied soft-clauses.
Leaf tables only hold a cardinality value of~$0$ as in Line~\ref{line:msat:leaf2}.
Then, similar to the table algorithm~\algo{Sat'} (cf., Algorithm~\ref{alg:primdb}), 
when introducing a variable~$a$, we guess the truth value
and keep those rows, where local formula~$\varphi_t$ is satisfied.
Whenever a variable~$a$ is removed in a node~$t$,
we remove column~$a$ and group 
rows that have common values over columns~$\chi(t)$.
Thereby, the new cardinality~\emph{card} for each group
is the maximum among the values of~\emph{card}
including the number of satisfied local soft-clauses~$\psi'$ of 
the child node of~$t$ that are removed in~$\psi_t$ (due to removal of~$a$).
Finally, similar to Algorithm~\ref{alg:vc}, a join node sums up cardinalities of
two child rows containing the same assignment.

\subsubsection*{Problem~\IDS}

\begin{algorithm}[ht]
  \KwData{Node~$t$, bag $\chi(t)$, local graph~$G_t$,
    and a sequence $\langle \tab{1},\ldots \tab{\ell}\rangle$ of child tables.}
\KwResult{Table $\tab{t}.\hspace{-5em}$} \lIf(\hspace{-1em})
  {$\type(t) = \leaf$}{%
    $\tab{t} \eqdef \tuplecolor{\inputPredColor}{\{(\text{card}, 0)\}}$ \label{line:ids:leaf2}%
  }%
  \uElseIf{$\type(t) = \intr$, and
    $a\hspace{-0.1em}\in\hspace{-0.1em}\chi(t)$ is introduced}{ %
    \hspace{-1em}$\tab{t}\eqdef \dot\Pi_{\chi(t),\tuplecolor{\inputPredColor}{\bigcup_{u\in \chi(t)}\{d_u \leftarrow \bigvee_{\{u,v\}\in E(G_t)} d_u \vee v, \text{card} \leftarrow \text{card}\}}}(\tab{1} \bowtie_{\tuplecolor{\inputPredColor}{\bigwedge_{\{u,v\}\in E(G_t)} \neg u \vee \neg v}} \tuplecolor{\inputPredColor}{\{\{(a, 0), (d_a, 0)\}, \{(a, 1), (d_a, 1)\}\}})\hspace{-10em}$ \label{line:ids:intr2}
    \vspace{-0.05em}
  }\vspace{-0.05em}%
  \uElseIf{$\type(t) = \rem$, and $a \not\in \chi(t)$ is removed}{%
    \hspace{-1em}$\tab{t} \eqdef {_{\chi(t)\cup\tuplecolor{\inputPredColor}{\{d_u\mid u\in\chi(t)\}}}G}_{\tuplecolor{\inputPredColor}{\text{card}\,\leftarrow\, \tab{}\mapsto\text{\ttfamily MIN}(\{r.\text{card} + r.a\mid r\in\tab{}\})}}(\Pi_{\attr(\tab{1})\setminus\{a,\tuplecolor{\inputPredColor}{\cid{d_a}}\}} \sigma_{\tuplecolor{\inputPredColor}{d_a}}\tab{1})$ \label{line:ids:rem2}\hspace{-5em} \vspace{-0.1em}
  } %
  \uElseIf{$\type(t) = \join$}{%
    \hspace{-1em}\makebox[3.3cm][l]{$\tab{t} \eqdef \dot\Pi_{\chi(t),\tuplecolor{\inputPredColor}{{\bigcup_{u\in\chi(t)}\{d_u \leftarrow d_u \vee d_u', \text{card} \leftarrow \text{card} + \text{card}'\}}}}(\tab{1} \bowtie_{\tuplecolor{\inputPredColor}{}{\bigwedge_{u\in\chi(t)}\cid{u}= \cid{u}'}} \rho_{\hspace{-0.75em}\bigcup\limits_{u\in \attr(\tab{2})}\hspace{-0.75em}\{\cid{u}\mapsto \cid{u}'\}}\tab{2})$} \label{line:ids:join2} %
    \hspace{-5em}%
    \vspace{-0.75em}
  }
  \Return $\tab{t}$ \vspace{-0.25em}
  \caption{Table algorithm~$\algo{IDS}(t,\chi(t),G_t,\langle \tab{1}, \ldots, \tab{\ell}\rangle)$ for solving~$\IDS$.}
  \label{alg:ids}
\end{algorithm}%
Given a graph instance~$\mathcal{I}=G=(V,E)$, a \emph{dominating set} of~$G$ is a set of vertices~$D\subseteq V$ of~$G$,
where each vertex~$v\in V$ is either in~$D$ or is adjacent some vertex in~$D$,~i.e., there is a vertex~$d\in D$ with~$\{d,v\}\in E$.
A dominating set~$D$ is an \emph{independent dominating set} of~$G$,
if there is no edge in~$E$ between vertices of~$D$.
Then, the problem \IDS asks to find the minimum cardinality~$\Card{D}$ among all independent dominating sets~$D$ of~$G$ (if exists). 
We define local instance by~$\IDS(t,G)\eqdef G_t$. 

The table algorithm for solving \IDS is given in Algorithm~\ref{alg:ids},
where a table~$\tab{t}$ of a node~$t$ uses 
column~\emph{card} for cardinalities of potential dominating sets,
and an additional Boolean column~$d_u$ per bag vertex~$u\in\chi(t)$.
Intuitively, $d_u$ indicates whether vertex~$u$ is already ``dominated'',
i.e., either~$u$ is in the dominating set or~$u$ has an adjacent vertex to the dominating set.
Similar to before, leaf nodes set cardinality \emph{card} to~$0$, cf., Line~\ref{line:ids:leaf2}.
For a node~$t$ with an introduced vertex~$a$, we guess in Line~\ref{line:ids:intr2}
whether~$a$ shall be in the dominating set or not (and set~$d_a$ to~$0$ or~$1$, respectively).
Then, %
we only keep rows that are independent, i.e., $a$ can not
be in the dominating set and adjacent to~$u$ in edges~$E(G_t)$ of local graph~$G_t$ at the same time.
Finally, %
values~$d_u$ (dominance status) for~$a$ and for neighbors of~$a$ are updated accordingly.
When a vertex~$a$ is removed in a node~$t$, Line~\ref{line:ids:rem2} only
keeps rows, where~$d_a$ is true, i.e., $a$ is indeed dominated, 
and removes columns~$a,d_a$.
Further, we group rows according to their values to~$\chi(t)\cup\{d_u\mid u\in\chi(t)\}$
and for each group we set the cardinality to the minimum among the cardinalities
of the group rows (including~$a$ if~$a$ is in the set).
For join nodes~$t$, Line~\ref{line:ids:join2} sums up cardinalities of rows holding the same dominating set and treats a vertex~$u\in\chi(t)$ as dominated if it is dominated in at least one of the two rows.

\medskip
Similar to \VC and \cTCOL one can model several other (graph) problems.
One could also think of counting the number of solutions to problem \VC,
where both a column for cardinalities and one for counting is used. 
There, in addition to grouping, we additionally could use conditions over groups
where only rows are kept whose column values for \emph{card} 
form the minimum within the group.

\subsection{Generalizing the Patterns of Table Algorithms}

\renewcommand*{\algorithmcfname}{\algorithmcfnameold}
\renewcommand*{\algorithmcfname}{\algorithmcfnamenew}
\begin{algorithm}[tb]
  \KwData{Node~$t$, bag $\chi(t)$, local instance~$\mathcal{I}_t$,
    and a sequence $\langle \tab{1},\ldots \tab{\ell}\rangle$ of child tables.}
\KwResult{Table $\tab{t}.\hspace{-5em}$} \lIf(\hspace{-1em})
  {$\type(t) = \leaf$}{%
    $\tab{t} \eqdef \tuplecolor{\inputPredColor}{\#\mathsf{leafTab}\#}$ \label{line:leaf2}%
  }%
  \uElseIf{$\type(t) = \intr$, and
    $a\hspace{-0.1em}\in\hspace{-0.1em}\chi(t)$ is introduced}{ %
    $\tab{t}\eqdef \dot\Pi_{\chi(t),\tuplecolor{\inputPredColor}{\#\mathsf{intrAddCols}\#}}(\tab{1} \bowtie_{\tuplecolor{\inputPredColor}{\#\mathsf{intrFilter}\#}} \tuplecolor{\inputPredColor}{\#\mathsf{intrTab}\#})$ \label{line:intr2}
    \hspace{-5em}\vspace{-0.05em}
  }\vspace{-0.05em}%
  \uElseIf{$\type(t) = \rem$, and $a \not\in \chi(t)$ is removed}{%
    $\tab{t} \eqdef {_{\chi(t)\cup\tuplecolor{\inputPredColor}{\#\mathsf{remGroupCols}\#}}G}_{\tuplecolor{\inputPredColor}{\#\mathsf{remAggr}\#}}(\Pi_{\attr(\tab{1})\setminus\{a,\tuplecolor{\inputPredColor}{\#\mathsf{remCols}\#}\}} \sigma_{\tuplecolor{\inputPredColor}{\#\mathsf{remFilter}\#}}\tab{1})$ \label{line:rem2}\hspace{-5em} \vspace{-0.1em}
  } %
  \uElseIf{$\type(t) = \join$}{%
    \makebox[3.3cm][l]{$\tab{t} \eqdef \dot\Pi_{\chi(t),\tuplecolor{\inputPredColor}{\#\mathsf{joinAddCols}\#}}(\tab{1} \bowtie_{\bigwedge_{u\in\chi(t)}u=u' \wedge {\tuplecolor{\inputPredColor}{\#\mathsf{joinAddFilter}\#}}} \rho_{\hspace{-0.75em}\bigcup\limits_{u\in \attr(\tab{2})}\hspace{-0.75em}\{\cid{u}\mapsto \cid{u}'\}}\tab{2})$} \label{line:join2} %
    \hspace{-75em}%
    \vspace{-0.75em}
  }
  \Return $\tab{t}$ \vspace{-0.25em}
  \caption{Template table algorithm~$\algo{A}(t,\chi(t),\mathcal{I}_t,\langle \tab{1}, \ldots, \tab{\ell}\rangle)$ %
    for solving problem~$\mathcal{P}$.}
  \label{alg:template}
\end{algorithm}%
\renewcommand*{\algorithmcfname}{\algorithmcfnameold}

  In the previous sections, we presented the table algorithms for
  solving a selection of combinatorial problems, namely, \cSAT,
  \cTCOL, \VC, \MSAT, and \IDS, by dynamic programming. As mentioned
  in Section~\ref{sec:prelimns:tds}, there are a variety of
  application areas where such algorithms allow for solving problems
  efficiently.
  When specifying most algorithms, we focus on the table
  algorithm~$\algo{A}$, which is executed for each node~$t$
  of~$T$ of the considered tree
  decomposition~$\mathcal{T}=(T,\chi)$ and computes a new table
  depending on the previously computed tables at the children of~$t$.
  From the descriptions above, it is easy to see that the algorithms
  effectively follow standard patterns. 
  Therefore, we present a general template in
  \algorithmcfnamenew~\ref{alg:template}, where parts of table
  algorithms for problems that are typically problem-specific are
  replaced by colored placeholders of the
  form~$\textcolor{\inputPredColor}{\#\mathsf{placeHolder}\#}$, cf.,
  \algorithmcfname~\ref{alg:primdb}.
  The general template of table algorithms works for many problems,
  including decision problems, counting problems as well as
  optimization problems.

  The intuition behind these placeholders is as follows: %
For leaf nodes, the initial table (typically empty) can be specified using~$\textcolor{\inputPredColor}{\#\mathsf{leafTab}\#}$.
For introduce nodes, the potential cases for the introduced vertex~$a$
are given with the help of~$\textcolor{\inputPredColor}{\#\mathsf{intrTab}\#}$.
Then, according to the local instance, we only keep those rows that
satisfy~$\textcolor{\inputPredColor}{\#\mathsf{intrFilter}\#}$.
The
placeholder~$\textcolor{\inputPredColor}{\#\mathsf{intrAddCols}\#}$
allows to add additional columns, which we often need when solving
problems that involve counting or optimizing a value.
In other words, placeholder~$\textcolor{\inputPredColor}{\#\mathsf{intrAddCols}\#}$ in Line~\ref{line:intr2} of \algorithmcfnamenew~\ref{alg:template}
uses extended projection, which %
is needed for problems requiring changes on vertex introduction.
Nodes, where an atom~$a$ is removed sometimes require to filter rows, which
do not lead to a solution using~$\textcolor{\inputPredColor}{\#\mathsf{remFilter}\#}$,
and to remove columns concerning~$a$ by~$\textcolor{\inputPredColor}{\#\mathsf{remCols}\#}$. Further, one oftentimes needs to aggregate rows
according to the values of the columns of the bag and additional columns (given by~$\textcolor{\inputPredColor}{\#\mathsf{remGroupCols}\#}$), where the aggregation is specified by~$\textcolor{\inputPredColor}{\#\mathsf{remAggr}\#}$.
Finally, for join nodes, one can specify an additional filter~$\textcolor{\inputPredColor}{\#\mathsf{joinAddFilter}\#}$ that goes beyond checking
equivalence of row values in the $\theta$-join operation. 
Further, depending on the problem one might need to add and update
the values of additional columns by using extended projection in form of placeholder~$\textcolor{\inputPredColor}{\#\mathsf{joinAddCols}\#}$.

Note that while the algorithms presented here assume for simplicity nice tree decompositions,
the whole architecture does not depend 
on certain restrictions of TDs, or whether it is nice or not.
Instead, a table algorithm of any TD is simply specified by 
handling \emph{problem-specific} implementations of the placeholders of \algorithmcfnamenew~\ref{alg:template}, where the system following this architecture is responsible for interleaving and overlapping these cases within a node~$t$.
In fact, we discuss an implementation of a system according to this architecture next, where it is crucial to implement non-nice TDs to obtain higher efficiency.

\section{System~\protect\dpdb: Dynamic Programming with Databases \& SQL}

\begin{figure*}[t]%
\centering%
  {\noindent\includegraphics[scale=0.9]{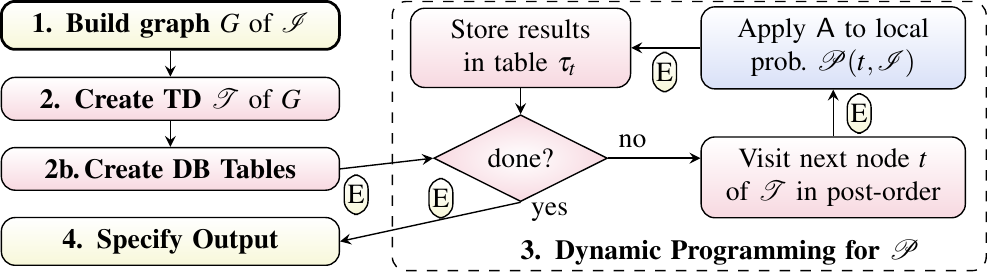}}%
  \caption{Architecture of Dynamic Programming with Databases. Steps highlighted in red are provided by the system depending on the specification of yellow and blue parts, which is given by the user for specific problems~$\mathcal{P}$. The yellow ``E''s represent events that can be intercepted and handled by the user. 
  The blue part concentrates on table algorithm~$\algo{A}$, where the user specifies how SQL code~is~generated in a modular~way.}
  \label{fig:arch}
\end{figure*}

In this section, we present a general architecture to model table algorithms
by means of database management systems.
We move from relational algebra definitions to specifications of the
table algorithms in terms of SQL queries.
The overall architecture is follows the DP approach as presented in Section~\ref{sec:dp}.
It works as depicted in Figure~\ref{fig:arch}
where the steps highlighted in yellow and blue need to be specified
depending on the problem~$\mathcal{P}$. Steps outside Step~3 are mainly setup tasks,
the yellow ``E''s indicate \emph{events} that might be needed to solve more complex problems
on the polynomial hierarchy. 
For example, one could create and drop auxiliary sub-tables for each node during Step~3 within such events.
Observe that after the generation of a TD~$\mathcal{T}=(T,\chi)$, 
Step~2b automatically creates tables~$\tab{t}$ for each node~$t$ of~$T$,
where the corresponding table columns of~$\tab{t}$ are specified in the blue part, i.e., 
within~$\algo{A}$. 
The \emph{default columns} of such a table~$\tab{t}$ that are assumed in this section foresee one column for each element of the bag~$\chi(t)$, where additional columns that are needed for solving the problem can be added. This includes additional auxiliary columns, which can be also counters or costs for counting or optimization, respectively.
Besides the definition of table schemes, the blue part concerns the specification of the table algorithm by means of a procedural \emph{generator template} that describes 
how to obtain SQL code %
for each node~$t$, thereby depending on~$\chi(t)$ and on the tables for child nodes of~$t$.
This generated SQL code is then used internally for manipulation of 
tables~$\tab{t}$ during the tree decomposition 
traversal in Step~3 of dynamic programming.

We implemented the proposed architecture of the previous section in the prototypical system \dpdb.
The system is open-source\footnote{Our system~\dpdb is available under GPL3 license
    at~\href{https://github.com/hmarkus/dp_on_dbs/releases/tag/1.1}{\nolinkurl{github.com/hmarkus/dp_on_dbs}}.}, written in Python 3, and uses PostgreSQL as DBMS.
We are certain that one can easily replace PostgreSQL by any other state-of-the-art relational database that uses SQL.
In the following, we discuss implementation specifics that are crucial for a performant system that is still extendable and flexible.

\paragraph*{Computing TDs.}
TDs are computed mainly with the library \emph{htd} version~1.2 with default
settings~\cite{AbseherMusliuWoltran17a}, %
which finds TDs extremely quick
also for interesting instances~\cite{FichteHecherZisser19a} due to heuristics.
Note that \dpdb directly supports the TD format of recent competitions~\cite{DellKomusiewiczTalmon18a},
i.e., one could easily replace the TD library.
It is important not to enforce htd to compute nice TDs, as this would cause a lot of overhead later in \dpdb for copying tables.
However, in order to benefit from the implementation of $\theta$-joins,
query optimization, and state-of-the-art database technology in general, 
we observed that it is crucial to limit the number of child nodes of every TD node.
In result, when huge tables are involved, $\theta$-joins among child node tables
cover at most a limited number of child node tables.
Hence, the query optimizer of the database system  can still come up with meaningful execution plans depending on the contents of the table.
Nonetheless we prefer $\theta$-joins with more than just two tables,
since binary $\theta$-joins already fix in which order these tables shall be combined, which already
 limits the query optimizer. 
Apart from this trade-off, we tried to outsource the task of joining tables to the DBMS, %
since the performance of database systems highly depends on query optimization. %
The actual limit, which is a restriction from experience and practice only, highly depends on the DBMS that is used.
For PostgreSQL, we set a limit of at most~$5$ child nodes for each node of the TD,~i.e., each $\theta$-join covers at most 5 child tables.

\paragraph*{Towards non-nice TDs.}
Although this paper presents the algorithms for nice TDs (mainly due to simplicity),
the system \dpdb interleaves these cases as presented in 
\algorithmcfname~\ref{alg:template}.
More precisely, the system executes one query per table~$\tab{t}$ for each node~$t$ during the traversal of TD~$\mathcal{T}$. 
This query consists of several parts and we briefly explain its parts from outside to inside in accordance with \algorithmcfname~\ref{alg:template}. 
First of all, the inner-most part concerns the \emph{row candidates} for~$\tab{t}$ consisting of the $\theta$-join among all child tables of~$\tab{t}$ as in Line~\ref{line:join} of \algorithmcfname~\ref{alg:template}. If there is no child node of~$t$, table $\tuplecolor{\inputPredColor}{\#\mathsf{leafTab}\#}$ of Line~\ref{line:leaf} is used instead.
Next, the result is cross-joined with $\tuplecolor{\inputPredColor}{\#\mathsf{intrTab}\#}$ for each introduced variable as in Line~\ref{line:intr}, but without using the filter~$\tuplecolor{\inputPredColor}{\#\mathsf{intrFilter}\#}$ yet. Then, the result is projected by using extended projection involving~$\chi(t)$ as well as both~$\tuplecolor{\inputPredColor}{\#\mathsf{joinAddCols}\#}$ and~$\tuplecolor{\inputPredColor}{\#\mathsf{intrAddCols}\#}$.
Actually, there are different configurations of how \dpdb can deal with the resulting row candidates.
For debugging (see below) one could (1) actually materialize the result in a table,
whereas for performance reasons, one should use (2) \emph{common table expressions (CTEs or {\ttfamily WITH}-queries)} or (3) \emph{sub-queries (nested queries)}, which both result in one nested SQL query per table~$\tab{t}$. 
On top of these row candidates, selection according to~$\tuplecolor{\inputPredColor}{\#\mathsf{intrFilter}\#}$, cf., Line~\ref{line:intr}, is executed.
Finally, the resulting table is plugged as table $\tab{1}$ into Line~\ref{line:rem},
where in particular the result is grouped by using both~$\chi(t)$\footnote{Actually, \dpdb keeps only columns relevant for the table of the parent node of~$t$.} and~$\tuplecolor{\inputPredColor}{\#\mathsf{remGroupCols}\#}$ and each group is aggregated by~$\tuplecolor{\inputPredColor}{\#\mathsf{remAggr}\#}$ accordingly.
It turns out that PostgreSQL can do better with sub-queries than CTEs, since we observed that the query optimizer
oftentimes pushes (parts of) outer selections and projections into the sub-query if needed, which
is not the case for CTEs, as discussed in the PostgreSQL manual~\cite[Sec. 7.8.1]{postgres}. On different DBMSs or other vendors, e.g., Oracle, it might be better to use CTEs instead.

\begin{example}\label{ex:dbviews}
Consider again Example~\ref{ex:running0} and Figure~\ref{fig:graph-td} and 
let us use table algorithm~$\mathsf{Sat'}$ with \dpdb on formula~$\varphi$ of TD~$\mathcal{T}$ and Option (3): sub-queries, where the row candidates are expressed via a sub-queries. Then, for each node~$t_i$ of~$\mathcal{T}$, \dpdb generates a view~$vi$ 
as well as a table~$\tab{i}$ containing in the end the content of~$vi$.
Observe that each view only has one column~$\cid{a}$ for each variable~$a$ of~$\varphi$ since the
truth assignments of the other variables are not needed later.
This keeps the tables compact, only $\tab{1}$ has two rows, $\tab{2}$, and $\tab{3}$ have only one row.
We obtain the following views.
\begin{alltt}\small
CREATE VIEW v1 AS SELECT a, sum(cnt) AS cnt FROM 
 (WITH intrTab AS (SELECT 0 AS val UNION ALL SELECT 1)
   SELECT i1.val AS a, i2.val AS b, i3.val AS c, 1 AS cnt 
          FROM intrTab i1, intrTab i2, intrTab i3)  
WHERE (NOT a OR b OR c) AND (a OR NOT b OR NOT c) GROUP BY a

CREATE VIEW v2 AS SELECT a, sum(cnt) AS cnt FROM 
 (WITH intrTab AS (SELECT 0 AS val UNION ALL SELECT 1) 
   SELECT i1.val AS a, i2.val AS d, 1 AS cnt FROM intrTab i1, intrTab i2) 
WHERE (a OR d) AND (a OR NOT d) GROUP BY a

CREATE VIEW v3 AS SELECT a, sum(cnt) AS cnt FROM 
 (SELECT \ensuremath{\tab{1}}.a, \ensuremath{\tab{1}}.cnt * \ensuremath{\tab{2}}.cnt AS cnt FROM \ensuremath{\tab{1}}, \ensuremath{\tab{2}} WHERE \ensuremath{\tab{1}}.a = \ensuremath{\tab{2}}.a)
GROUP BY a\end{alltt}%
\end{example}%

\paragraph*{Parallelization.} A further reason to not over-restrict the number of child nodes within the TD, lies in parallelization.
In \dpdb, we compute tables in parallel along the TD,
where multiple tables can be computed at the same time,
as long as the child tables are computed.
Therefore, we tried to keep the number of child nodes in the TD as high as possible.
In our system \dpdb, we currently allow 
for at most 24 worker threads for table computations and 24 database connections at the same time (both pooled and configurable).
On top of that we have 2 additional threads and database connections for job assignments to workers, as well as one dedicated watcher thread for clean-up and connection termination, respectively.

\paragraph*{Logging, Debugging and Extensions.} Currently, we have two versions of the \dpdb system implemented.
One version aims for performance and the other one tries to achieve comprehensive logging and easy debugging of problem (instances), thereby increasing explainability.
The former  does neither keep intermediate results 
nor create database tables in advance (Step 2b),
as depicted in Figure~\ref{fig:arch}, but creates tables according 
to an SQL {\ttfamily SELECT} statement.
In the latter, we keep all intermediate results, we record database timestamps before and after certain nodes, provide statistics as,~e.g., width and number of rows.
Further, since for each table~$\tab{t}$, exactly one SQL statement is executed for filling this table, we also have a dedicated view of the SQL {\ttfamily SELECT} statement, whose result is then inserted in~$\tab{t}$.
Together with the power and flexibility of SQL queries, we observed that this helps in finding errors in the table algorithm specifications.

Besides convient debugging, system \dpdb immediately
contains an extension for \emph{approximation}.
There, we restrict the table contents to a maximum number of rows.
This allows for certain approximations on counting problems or
optimization problems, where it is infeasible to compute the full tables.
Further, \dpdb foresees a dedicated \emph{randomization} on these restricted number of rows
such that we obtain different approximate results on different random seeds.

Note that \dpdb can be easily extended. 
Each problem can overwrite existing default behavior and \dpdb also supports
problem-specific argument parsers for each problem individually.
Out-of-the-box, we support the formats DIMACS SAT and DIMACS graph~\cite{LiuZhongJiao06} as well as the common format for TDs~\cite{DellKomusiewiczTalmon18a}.

\subsection*{Implementing table algorithms with~\protect\dpdb for selected problems}

The system \dpdb allows for \emph{easy prototyping} of DP algorithms on TDs.
In the following, we present the relevant parts of table algorithm implementations
according to the template in \algorithmcfname~\ref{alg:template} for our selection of problems below\footnote{Prototypical implementations for problems~\cSAT as well as~\VC are readily available in~\dpdb.}.
More precisely, we give the SQL implementations of the table algorithms of the previous section in form of specifying the corresponding placeholders as given by the template algorithm~$\algo{A}$.
Thereby, we only specify the placeholders needed for solving the problems,~i.e., placeholders of template algorithm~$\algo{A}$ that are not used (empty) are left out.
To this end, we assume in this section for each problem a not necessarily nice TD~$\mathcal{T}=(T,\chi)$ of the corresponding graph representation of our given instance~$\mathcal{I}$,
as well as
any node~$t$ of~$T$ and its child nodes~$t_1, \ldots, t_\ell$.

\paragraph*{Problem~$\cSAT$.}
Given instance formula~$\mathcal{I}=\varphi$.
Then, the specific placeholders of the template for~$\cSAT$ for a node~$t$ with $\varphi_t = \{\{l_{1,1},\ldots,l_{1,k_1}\}, \ldots, \{l_{n,1},\ldots,l_{n,k_n}\}\}$
that are required for \dpdb to solve the problem are as follows.
\begin{itemize}
	\item\makebox[8.25em][l]{\tuplecolor{\inputPredColor}{$\#\mathsf{leafTab}\#$}:}{\ttfamily SELECT 1 AS cnt}
	\item\makebox[8.25em][l]{\tuplecolor{\inputPredColor}{$\#\mathsf{intrTab}\#$}:}{\ttfamily SELECT 0 AS val UNION ALL SELECT 1}
	\item\makebox[8.25em][l]{\tuplecolor{\inputPredColor}{$\#\mathsf{intrFilter}\#$}:}{\ttfamily $(l_{1,1}$ OR $\ldots$ OR $l_{1,k_1})$ AND $\ldots$ AND $(l_{n,1}$ OR $\ldots$ OR $l_{n,k_n})$}
	\item\makebox[8.25em][l]{\tuplecolor{\inputPredColor}{$\#\mathsf{remAggr}\#$}:}{\ttfamily SUM(cnt) AS cnt}
	\item\makebox[8.25em][l]{\tuplecolor{\inputPredColor}{$\#\mathsf{joinAddCols}\#$}:}{\ttfamily \tab{1}.cnt * $\ldots$ * $\tab{\ell}$.cnt AS cnt }
\end{itemize}
If one compares this specification to Algorithm~\ref{alg:primdb}, one sees that conceptually the same idea is given above.
However, for efficiency \dpdb does not rely on nice TDs.
Observe that for the plain decision problem~$\SAT$, where the goal is to decide only the existence of a satisfying assignment for given formula~$\varphi$, 
placeholder $\tuplecolor{\inputPredColor}{\#\mathsf{leafTab}\#}$ would need to return the empty table and 
parts $\tuplecolor{\inputPredColor}{\#\mathsf{remAggr}\#}$ and $\tuplecolor{\inputPredColor}{\#\mathsf{joinAddCols}\#}$ are just empty since no counter~\emph{cnt} is needed.

\paragraph*{Problem~$\cTCOL$.}
Recall the problem~$\cTCOL$ and Algorithm~\ref{alg:col}.
Let~$\mathcal{I}=G=(V,E)$ be a given input graph.
Then, specific implementation parts for~$\cTCOL$ for a node~$t$ with~$E(G_t)=\{\{u_1,v_1\},\ldots,$ $\{u_n,v_n\}\}$ is given as follows.
\begin{itemize}
	\item\makebox[8.25em][l]{\tuplecolor{\inputPredColor}{$\#\mathsf{leafTab}\#$}:}{\ttfamily SELECT 1 AS cnt}
	\item\makebox[8.25em][l]{\tuplecolor{\inputPredColor}{$\#\mathsf{intrTab}\#$}:}{\ttfamily SELECT 0 AS val UNION ALL $\ldots$ UNION ALL SELECT $o$}
	\item\makebox[8.25em][l]{\tuplecolor{\inputPredColor}{$\#\mathsf{intrFilter}\#$}:}{\ttfamily NOT $(\cid{u_1}=\cid{v_1})$ AND $\ldots$ AND NOT $(\cid{u_n}=\cid{v_n})$}
	\item\makebox[8.25em][l]{\tuplecolor{\inputPredColor}{$\#\mathsf{remAggr}\#$}:}{\ttfamily SUM(cnt) AS cnt}
	\item\makebox[8.25em][l]{\tuplecolor{\inputPredColor}{$\#\mathsf{joinAddCols}\#$}:}{\ttfamily $\tab{1}$.cnt * $\ldots$ * $\tab{\ell}$.cnt AS cnt }
\end{itemize}

\paragraph*{Problem~\VC.}
Given any input graph~$\mathcal{I}=G=(V,E)$ of \VC.
Then, problem $\VC$ for a node~$t$ with~$E(G_t)=\{\{u_1,v_1\},\ldots, \{u_n,v_n\}\}$ and removed vertices~$\chi(t)\setminus(\chi(t_1) \cup \ldots \cup \chi(t_\ell))=\{r_1,\ldots,r_{m}\}$ is specified by the following placeholders (cf., Algorithm~\ref{alg:vc}).

\begin{itemize}
	\item\makebox[8.25em][l]{\tuplecolor{\inputPredColor}{$\#\mathsf{leafTab}\#$}:}{\ttfamily SELECT 0 AS card}
	\item\makebox[8.25em][l]{\tuplecolor{\inputPredColor}{$\#\mathsf{intrTab}\#$}:}{\ttfamily SELECT 0 AS val UNION ALL SELECT 1}
	\item\makebox[8.25em][l]{\tuplecolor{\inputPredColor}{$\#\mathsf{intrFilter}\#$}:}{\ttfamily $(\cid{u_1}$ OR $\cid{v_1})$ AND $\ldots$ AND $(\cid{u_n}$ OR $\cid{v_n})$}
	\item\makebox[8.25em][l]{\tuplecolor{\inputPredColor}{$\#\mathsf{remAggr}\#$}:}{\ttfamily MIN(card + $r_1$ + $\ldots$ + $r_m$) AS card}
	\item\makebox[8.25em][l]{\tuplecolor{\inputPredColor}{$\#\mathsf{joinAddCols}\#$}:}{\ttfamily $\tab{1}$.card + $\ldots$ + $\tab{\ell}$.card AS card}%
\end{itemize}

\paragraph*{Problem~\MSAT.} Given an instance~$\mathcal{I}=(\varphi, \psi)$ of problem~$\MSAT$. 
Then, the problem for a node~$t$ with local hard clauses~$\varphi_t = \{\{l_{1,1},\ldots,l_{1,k_1}\}, \ldots, \{l_{n,1},\ldots,l_{n,k_n}\}\}$ and local soft clauses~$\psi_t = \{\{l'_{1,1},\ldots,l'_{1,k'_1}\}, \ldots, \{l'_{p,1},\ldots,l_{p,k'_p}\}\}$ is specified by the following placeholders (cf., Algorithm~\ref{alg:msat}).
\begin{itemize}
	\item\makebox[8.25em][l]{\tuplecolor{\inputPredColor}{$\#\mathsf{leafTab}\#$}:}{\ttfamily SELECT 0 AS card}
	\item\makebox[8.25em][l]{\tuplecolor{\inputPredColor}{$\#\mathsf{intrTab}\#$}:}{\ttfamily SELECT 0 AS val UNION ALL SELECT 1}
	\item\makebox[8.25em][l]{\tuplecolor{\inputPredColor}{$\#\mathsf{intrFilter}\#$}:}{\ttfamily $(l_{1,1}$ OR $\ldots$ OR $l_{1,k_1})$ AND $\ldots$ AND $(l_{n,1}$ OR $\ldots$ OR $l_{n,k_n})$}
	\item\makebox[8.25em][l]{\tuplecolor{\inputPredColor}{$\#\mathsf{remAggr}\#$}:}{\ttfamily MIN(card + $(l'_{1,1}$ OR $\ldots$ OR $l'_{1,k'_1})$ + $\ldots$ + }%
	\item[] \makebox[14em][l]{}{\ttfamily $(l'_{p,1}$ OR $\ldots$ OR $l'_{p,k'_p})$) AS card}
	\item\makebox[8.25em][l]{\tuplecolor{\inputPredColor}{$\#\mathsf{joinAddCols}\#$}:}{\ttfamily $\tab{1}$.card + $\ldots$ + $\tab{\ell}$.card AS card}
\end{itemize}

\paragraph*{Problem~\IDS.} Recall an instance~$\mathcal{I}=G=(V,E)$ of problem~$\IDS$ and table algorithm~$\algo{IDS}$ as given in Algorithm~\ref{alg:ids}. 
The implementation of table algorithm~$\algo{IDS}$ for~$\IDS$ for a node~$t$ assumes that~$E(G_t)=\{\{u_1,v_1\},\ldots,$ $\{u_n,v_n\}\}$. Further, we let bag~$\chi(t)=\{a_1,\ldots,a_k\}$, removed vertices~$\chi(t)\setminus(\chi(t_1) \cup \ldots \cup \chi(t_\ell))=\{r_1,\ldots,r_{m}\}$, and we let the~$w_i$ many neighbors~$N_i$ of each vertex~$a_i$ in~$G$ (with $1\leq i\leq k$) be given by~$N_i=\{e_{a_i,1}, \ldots, e_{a_i,w_i}\}$. Then, the SQL implementation can be specified as follows.
\begin{itemize}
	\item\makebox[8.25em][l]{\tuplecolor{\inputPredColor}{$\#\mathsf{leafTab}\#$}:}{\ttfamily SELECT 0 AS card}
	\item\makebox[8.25em][l]{\tuplecolor{\inputPredColor}{$\#\mathsf{intrTab}\#$}:}{\ttfamily SELECT 0 AS val, 0 AS d UNION ALL SELECT 1, 1}
	\item\makebox[8.25em][l]{\tuplecolor{\inputPredColor}{$\#\mathsf{intrFilter}\#$}:}{\ttfamily $($NOT $\cid{u_1}$ OR NOT $\cid{v_1})$ AND $\ldots$ AND $($NOT $\cid{u_n}$ OR NOT $\cid{v_n})$}
	\item\makebox[8.25em][l]{\tuplecolor{\inputPredColor}{$\#\mathsf{intrAddCols}\#$}:}{\ttfamily card, $d_{a_1}$ OR $e_{a_1,1}$ OR $\ldots$ OR $e_{a_1,w_1}$ AS $d_{a_1}$, $\ldots$} %
\item[]\makebox[11.5em][l]{}{\ttfamily $d_{a_k}$ OR $e_{a_k,1}$ OR $\ldots$ OR $e_{a_k,w_k}$ AS $d_{a_k}$}
	\item\makebox[8.25em][l]{\tuplecolor{\inputPredColor}{$\#\mathsf{remFilter}\#$}:}{\ttfamily $d_{a_1}$ AND $\ldots$ AND $d_{a_k}$}
	\item\makebox[8.25em][l]{\tuplecolor{\inputPredColor}{$\#\mathsf{remAggr}\#$}:}{\ttfamily MIN(card  + $r_1$ + $\ldots$ + $r_m$) AS card}
	\item\makebox[8.25em][l]{\tuplecolor{\inputPredColor}{$\#\mathsf{remGroupCols}\#$}:}{\ttfamily $d_{a_1}$, $\ldots$, $d_{a_k}$}
	\item\makebox[8.25em][l]{\tuplecolor{\inputPredColor}{$\#\mathsf{joinAddCols}\#$}:}{\ttfamily $\tab{1}$.card + $\ldots$ + $\tab{\ell}$.card AS card,}%
	\item[]\makebox[8.25em][l]{}{\ttfamily $\tab{1}$.$d_{a_1}$ OR $\ldots$ OR $\tab{\ell}$.$d_{a_1}$ AS $d_{a_1}$, $\ldots$}
	\item[]\makebox[8.25em][l]{}{\ttfamily $\tab{1}$.$d_{a_k}$ OR $\ldots$ OR $\tab{\ell}$.$d_{a_k}$ AS $d_{a_k}$}
\end{itemize}

Note that implementations could generate and apply parts of this specification, 
as for example in \tuplecolor{\inputPredColor}{$\#\mathsf{intrFilter}\#$} only edges that have not been checked so far in any descending node, need to be checked. %

Similar to \VC, \cTCOL, %
and \IDS one can model several related (graph) problems.
One could also think of counting the number of solutions to problem \MSAT,
where both, a column for cardinalities and one for counting is used. 
There, in addition to grouping with {\ttfamily GROUP BY} in \dpdb, we additionally use the {\ttfamily HAVING} construct of SQL,
where only rows are kept, whose column {\ttfamily card} is minimal.

\section{Experiments}
\label{sec:experiments}
We conducted a series of experiments using publicly available benchmark sets for~\cSAT.
Our tested benchmarks~\cite{FichteEtAl18b} 
are publicly available 
and our results are
also on github at
\href{https://github.com/hmarkus/dp_on_dbs/tree/tplp}{\nolinkurl{github.com/hmarkus/dp_on_dbs/tplp}}.

\subsection{Setup}

\paragraph{Measure \& Resources.}
We mainly compare wall clock time and number of
timeouts. 
In the time we include \emph{preprocessing
  time} as well as \emph{decomposition time} for computing a %
TD with a fixed random seed. %
For parallel solvers we allowed access to 24 physical cores on
machines. %
We set a timeout of 900 seconds and limited available RAM to~14 GB per
instance and solver.
However, since our solver \dpdb is a solver using multiple threads, we 
 restricted the results of \dpdb to a total of 900 seconds of CPU time.
While allowing for all the other (parallel) solvers more than 900 seconds of CPU time.
For \dpdb, we only allowed 900 seconds of CPU time, but at the same time 
restricted to 900 seconds wall clock time.
\paragraph{Benchmark Instances.}
We considered a selection of overall 1494 instances from various
publicly available benchmark sets~\cSAT consisting of %
\instances{fre/meel} benchmarks\footnote{See:
  \href{http://tinyurl.com/countingbenchmarks}{\nolinkurl{tinyurl.com/countingbenchmarks}}}(1480
instances), %
and \instances{c2d} benchmarks\footnote{See:
  \href{http://reasoning.cs.ucla.edu/c2d/results.html}{\nolinkurl{reasoning.cs.ucla.edu/c2d}}}
(14 instances).
We  preprocessed the instances by the \cSAT preprocessor 
\emph{pmc}~\cite{LagniezMarquis14a}, similar to results of recent work on~\cSAT~\cite{FichteHecherZisser19a}, where it was also shown that more than 80\% of the \cSAT instances have primal treewidth below~19
after preprocessing.
For preprocessing with \emph{pmc} we used the recommended options \texttt{-vivification -eliminateLit -litImplied -iterate=10 -equiv -orGate -affine}, which ensures that model counts are preserved.

\paragraph{Benchmarked system \protect\dpdb.}
We used PostgreSQL 12 for our system~\dpdb on a tmpfs-ramdisk (/tmp) that could grow up to 
at most 1 GB per run.
To ensure comparability with previous results~\cite{FichteEtAl20},
where we had employed PostgreSQL 9.5 for our system~\dpdb, 
we also considered the configuration~\dpdbold that uses 
the preinstalled database system PostgreSQL 9.5
without any ramdisk at all (plain hard disk).
However, we observed major performance increases of \dpdb compared to \dpdbold.
  We allow parallel execution for the database management system
  PostgreSQL with access to all cores of the benchmarking
  system. However, we restrict the total CPU time to ensure that we do
  not bias system resources  towards dpdb.
\paragraph{Other benchmarked systems.}
In our experimental work, we present results for the most recent
versions of publicly available \cSAT solvers, namely,
\href{http://reasoning.cs.ucla.edu/c2d/download.php}{\textit{c2d}~2.20}~\cite{Darwiche04a},
\href{http://www.cril.univ-artois.fr/KC/d4.html}{\textit{d4}~1.0}~\cite{LagniezMarquis17a},
\href{https://bitbucket.org/haz/dsharp}{\textit{DSHARP}~1.0}~\cite{MuiseEtAl12a},
\href{http://reasoning.cs.ucla.edu/minic2d/}{\textit{miniC2D}~1.0.0}~\cite{OztokDarwiche15a},
\href{http://www.cril.univ-artois.fr/KC/eadt.html}{\textit{cnf2eadt}~1.0}~\cite{KoricheLagniezMarquisThomas13a}, 
\href{http://www.sd.is.uec.ac.jp/toda/code/cnf2obdd.html}{\textit{bdd\_{}minisat}~1.0.2}~\cite{TodaSoh15a},
and \href{http://reasoning.cs.ucla.edu/sdd/}{\textit{sdd}~2.0}~\cite{Darwiche11a}, which are all based on %
knowledge compilation techniques.
\begin{figure}[t]
  \centering
  \resizebox{1\columnwidth}{!}{\includegraphics{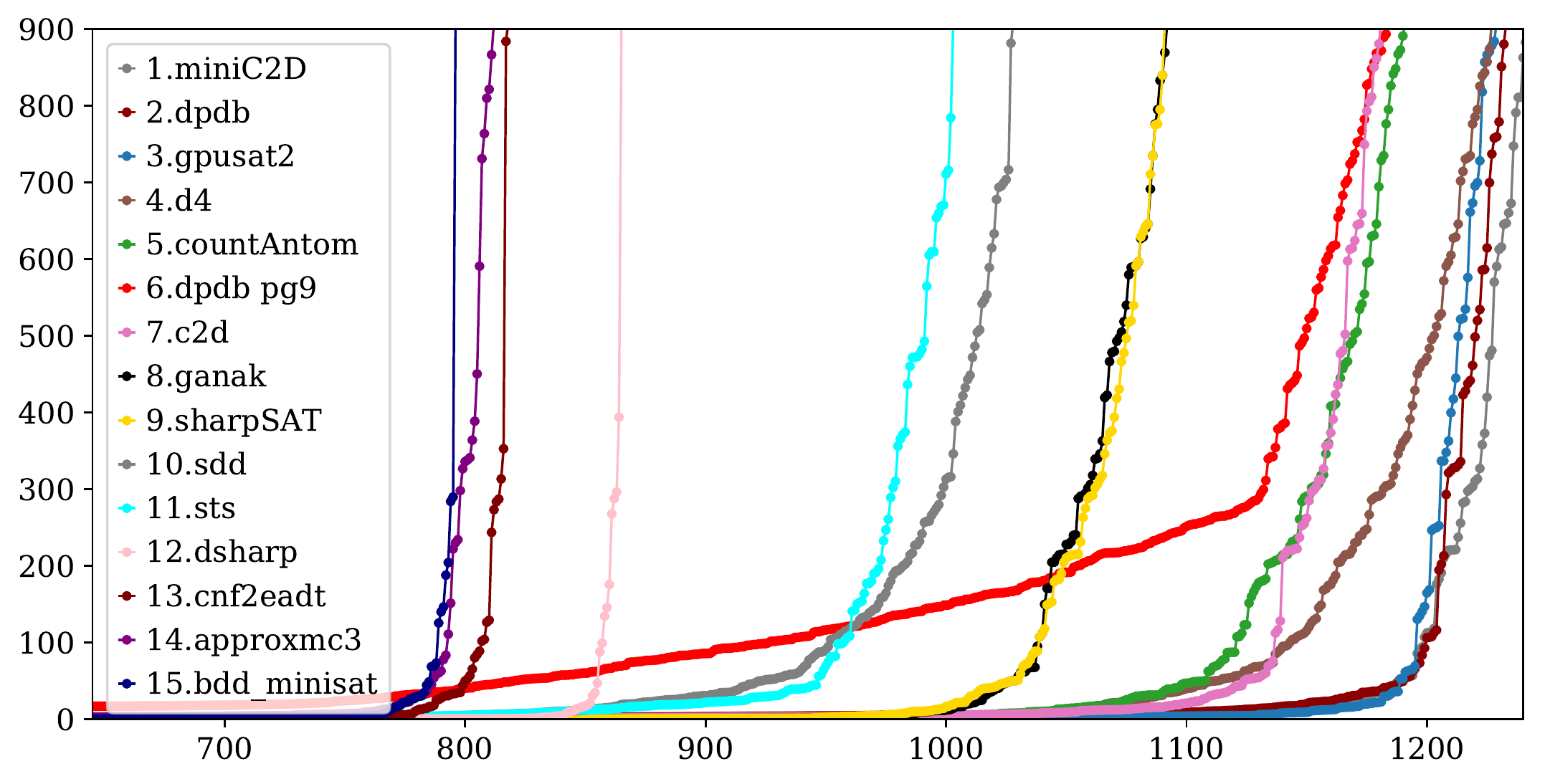}}
  \caption{Runtime for the top 15 solvers over {all}~\cSAT instances.
    The x-axis refers to the number of instances and the y-axis
    depicts the runtime sorted in ascending order for each solver
    individually.}
  \label{fig:runtime}
\end{figure}
We also considered rather recent approximate solvers
\href{https://bitbucket.org/kuldeepmeel/approxmc}{\textit{ApproxMC2}, \textit{ApproxMC3}}~\cite{ChakrabortyEtAl14a}, 
{\href{https://github.com/meelgroup/ganak}{\textit{ganak}}}~\cite{SharmaEtAl19}, and
\href{http://cs.stanford.edu/~ermon/code/STS.zip}{\textit{sts}~1.0}~\cite{ErmonGomesSelman12a},
as well as %
CDCL-based solvers
\href{https://www.cs.rochester.edu/u/kautz/Cachet/cachet-wmc-1-21.zip}{\textit{Cachet}~1.21}~\cite{SangEtAl04},
\href{http://tools.computational-logic.org/content/sharpCDCL.php}{\textit{sharpCDCL}}\footnote{See:
  \href{http://tools.computational-logic.
    org/content/sharpCDCL.php}{\nolinkurl{tools.computational-logic.
      org}}}, %
and
\href{https://sites.google.com/site/marcthurley/sharpsat}{\textit{sharpSAT}~13.02}~\cite{Thurley06a}.
Finally, we also included multi-core solvers
\href{https://github.com/daajoe/GPUSAT/releases/tag/v0.815-pre}{\textit{gpusat}~1.0 and \textit{gpusat}~2.0}~\cite{FichteEtAl18c,FichteHecherZisser19a}, which both are based on dynamic programming, as well as
\textit{countAntom}~1.0~\cite{BurchardSchubertBecker15a} on 12 physical CPU
cores, which performed better than on 24 cores.
Experiments were conducted with default~solver~options.
  Note that we excluded distributed solvers such as
  dCountAntom~\cite{BurchardSchubertBecker16a} and
  DMC~\cite{LagniezMarquisSzczepanski18a} from our experimental
  setup. Both solvers require a cluster with access to the Open-MPI
  framework~\cite{GabrielFaggBosilca04} and fast physical
  interconnections. Unfortunately, we do not have access to OpenMPI on
  our cluster. Nonetheless, our focus are shared-memory systems and
  since dpdb might well be used in a distributed setting, it leaves an
  experimental comparison between distributed solvers that also
  include dpdb as subsolver to future work.
\paragraph{Benchmark Hardware.}
Almost all solvers were executed on a cluster of 12 nodes. Each node is equipped
with two Intel Xeon E5-2650 CPUs consisting of 12 physical cores each
at 2.2 GHz clock speed, 256 GB RAM and 1 TB hard disc drives (\emph{not} an SSD) Seagate ST1000NM0033. %
The results were gathered on Ubuntu~16.04.1 LTS machines with disabled hyperthreading
on kernel~4.4.0-139. %
As we also took into account solvers using a GPU, 
for~\gpusatone and~\gpusatnu we used a machine equipped with a consumer GPU:
Intel Core i3-3245 CPU operating at 3.4 GHz, 16 GB RAM, and one
Sapphire Pulse ITX Radeon RX 570 GPU running at 1.24 GHz with 32
compute units, 2048 shader units, and 4GB VRAM using driver
amdgpu-pro-18.30-641594 and OpenCL~1.2.
The system operated on Ubuntu~18.04.1 LTS with kernel 4.15.0-34.

\newcommand{\inacc}[1]{\ensuremath{\diamond{}}#1}
\begin{table}[tb]
  \centering
  \resizebox{1\columnwidth}{!}{%
    \includegraphics{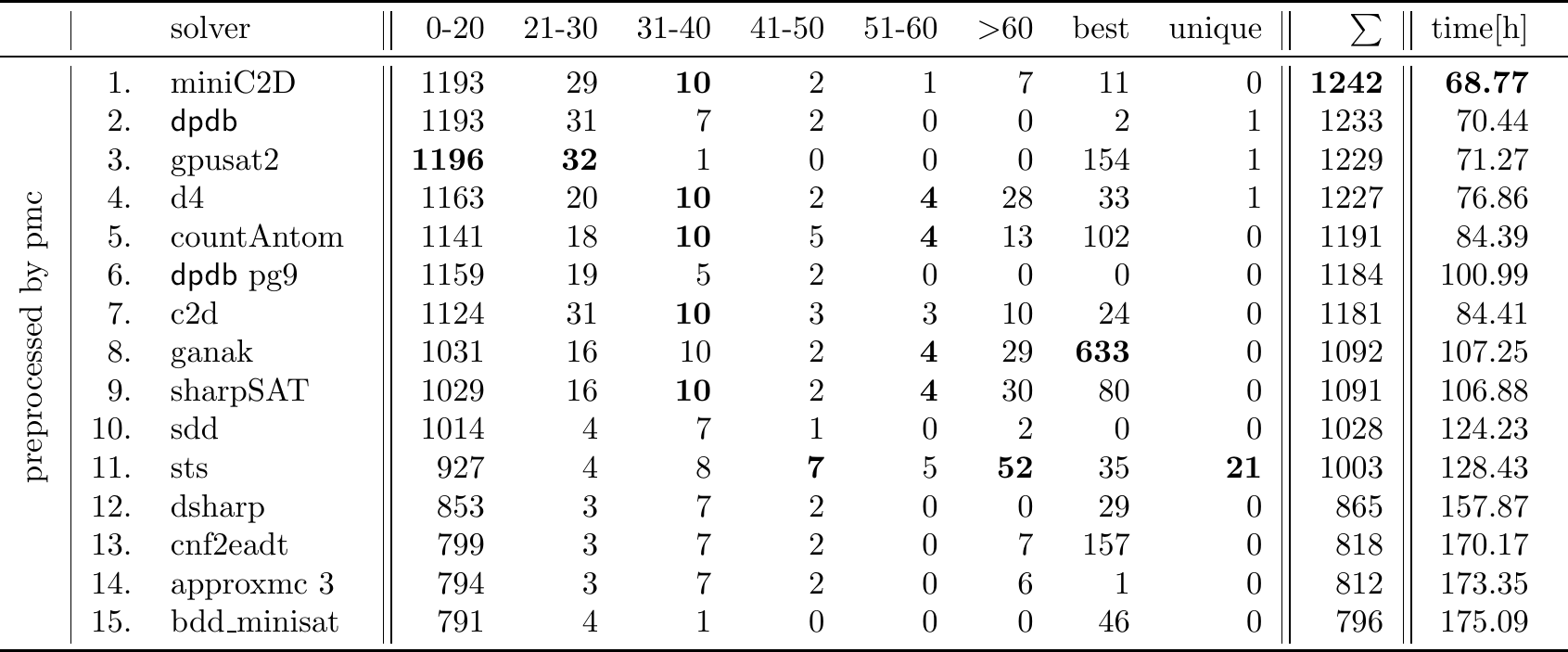}
  }
  \caption{%
    Number of solved~\cSAT instances, preprocessed by pmc and grouped by intervals of upper bounds of the treewidth.
    time[h] is the cumulated  wall clock time in hours, where unsolved instances 
    are counted as 900 seconds.
  }%
  \label{tab:sat:merged}
\end{table}%

\subsection{Results} %

Figure~\ref{fig:runtime} illustrates the top 15 solvers,
where instances are preprocessed by pmc, %
in a cactus-like plot, which provides an overview over all the benchmarked~\cSAT instances.
 The x-axis of these plots refers to the number 
of instances and the y-axis depicts the runtime sorted in ascending order for each solver individually.
Overall, \dpdb seems to be quite competitive and beats most of the solvers, as for example d4, countAntom, c2d, ganak, sharpSAT, dsharp, and approxmc.
Interestingly, \dpdb solves also instances, whose treewidth upper bounds are larger than~41.
Surprisingly, \dpdbold shows a different runtime behavior than the other solvers. 
We believe that the reason lies in an initial overhead caused by the creation of the tables that seems to depend on the number of nodes of the used TD.
There, \emph{I/O operations} of writing from main memory to hard disk seem to kick in.
This disadvantage is resolved if benchmarking \dpdb on recent versions of PostgreSQL (version~12) and using tmpfs-ramdisks instead of plain hard disks.
Table~\ref{tab:sat:merged} presents more detailed runtime results, showing a solid second place for \dpdb as our system solves the vast majority of the instances. %
Notably, it seems that the ganak solver is among the fastest on a lot of instances.
We observed that ganak has the fastest runtime on 633 instances,
when considering results of all~15 presented solvers.
Assume we only have instances up to an upper bound\footnote{These upper bounds were obtained via decomposer htd in at most two seconds.} of treewidth~35.
Then, if only instances with TDs up to width~50 are considered, \dpdb solves about the same number of instances than miniC2D solves. %

\section{Final Discussion \& Conclusions}
We presented a generic system~\dpdb for explicitly exploiting treewidth
by means of dynamic programming on databases.
The idea of~\dpdb is to use database management systems (DBMSs)
for table manipulation, which makes it (i)~easy and elegant to
perform \emph{rapid prototyping} for problems with DP algorithms 
and (ii)~allows to leverage  decades of database theory and 
database system tuning.
It turned out that all the cases that occur in dynamic programming can be handled
quite elegantly with plain SQL queries.
Our system~\dpdb can be used for both decision and counting problems,
thereby also considering optimization.
We see our system particularly well-suited for counting problems,
especially, since it was shown that for model counting (\cSAT)
instances of practical relevance typically have small treewidth~\cite{FichteHecherZisser19a}.
In consequence, we carried out preliminary experiments on publicly available instances for~\cSAT, 
where we see competitive behavior compared to most recent solvers.
\subsection*{Future Work}
Our results give rise to several research questions.
We want to push towards other database systems and vendors. For example, 
 we expect major improvements in commercial database management systems due to the 
availability of efficient enterprise features.
In particular, we expect in the DBMS Oracle that the behavior
when we use different strategies on how to write and evaluate our SQL queries,~e.g., 
sub-queries compared to common table expressions.
Currently, we do not create or use any indices,
as preliminary tests showed that \emph{meaningful B*tree indices} are hard to create
and creation is oftentimes too expensive. %
Further, the exploration of bitmap indices, as available in Oracle \emph{enterprise DBMS} would be worth trying in our case (and for~\cSAT), 
since one can efficiently combine database columns by 
using extremely \emph{efficient bit operations}.
It would also be interesting to investigate whether operating system
features to handle memory access can be
helpful~\cite{FichteMantheyStecklina20}.  In addition, one might
consider dpdb in the setting of distributed algorithms such as
dCountAntom~\cite{BurchardSchubertBecker16a} and
DMC~\cite{LagniezMarquisSzczepanski18a}.

It might be worth to rigorously test and explore our extensions on limiting the number of rows per table for \emph{approximating} \cSAT or other counting problems, cf.,~\cite{ChakrabortyMeelVardi16a,MeelEtAl17a,SharmaEtAl19} and compare to the recent winners of the newly established model counting competition~\cite{FichteHecherHamiti20}.
Recent results~\cite{HecherThierWoltran20} indicate that by using
hybrid solving and abstractions our results can also be extended to
projected model counting~\cite{FichteEtAl18d}.

Another interesting research direction is to study whether
efficient data representation techniques on DBMSs can be combined with dynamic
programming in order to lift our solver to quantified Boolean formulas.

It would also be interested to consider other measures such as
(fractional) hypertree
width~\cite{FichteHecherSzeider20,DzulfikarFichteHecher19} and
investigate whether tree decompositions with additional
properties~\cite{JegouTerrioux14} or other heuristics to compute tree
decompositions improve solving~\cite{Strasser17}.  Furthermore, 
interesting directions for future research would be to implement
counting various problems in our framework, such as in constraint
satisfaction~\cite{DurandMengel15,KhamisNgoRudra16}, constraint
networks~\cite{JegouTerrioux14},
argumentation~\cite{FichteHecherMeier19}, description
logics~\cite{FichteHecherMeier21}, or epistemic logic
programming~\cite{HecherMorakWoltran20}.

\section*{System and License}
Our system~\dpdb is available under GPL3 license
at~\href{https://github.com/hmarkus/dp_on_dbs/releases/tag/1.1}{\nolinkurl{github.com/hmarkus/dp_on_dbs}}.
\section*{Acknowledgements} 
    Part of this work was done while Johannes K. Fichte was visiting
    the Simons Institute for the Theory of Computing. 
    Main work was carried out while he was a PostDoc at TU Dresden.
    The work has been supported by the Austrian Science Fund (FWF),
    Grants Y698 and P32830, as well as the Vienna Science and Technology Fund, Grant WWTF ICT19-065.
    Markus Hecher is also affiliated with the University of Potsdam, Germany.
\bibliographystyle{acmtrans}
\bibliography{dpdb-tplp20}

\end{document}